\documentclass{article}

\PassOptionsToPackage{numbers, compress}{natbib}

\usepackage[final]{neurips_2024}

\usepackage[utf8]{inputenc} 
\usepackage[T1]{fontenc}    
\usepackage{hyperref}       
\usepackage{url}            
\usepackage{booktabs}       
\usepackage{amsfonts}       
\usepackage{nicefrac}       
\usepackage{microtype}      
\usepackage{xcolor}         

\usepackage{amsmath}
\usepackage{amssymb}
\usepackage{appendix}
\usepackage{tikz}
\usetikzlibrary{positioning}
\usetikzlibrary{shapes.geometric}

\usepackage{subfig}
\usepackage{booktabs}
\usepackage{xspace}
\usepackage{wrapfig}
\usepackage{adjustbox}
\usepackage{algorithm}
\usepackage{algpseudocode}
\usepackage{multicol}
\usepackage{multirow}
\usepackage{arydshln}
\usepackage{bbm}

\newcommand{\tabref}[1]{Table\xspace(\ref{#1})}

\newcommand{\IGNORE}[1]{}

\newcommand{\KL}{D_{\mathrm{KL}}}

\newcommand\sect[1]{\S\ref{#1}}

\newcommand{\appref}[1]{Appendix\xspace(\ref{#1})}

\def\tabref#1{Table~\ref{#1}}
\def\Tabref#1{Table~\ref{#1}}

\def\figref#1{Figure~\ref{#1}}
\def\Figref#1{Figure~\ref{#1}}

\def\secref#1{Section~\ref{#1}}

\def\eqref#1{equation~\ref{#1}}

\title{Beyond Internal Data: \\ Constructing Complete Datasets for Fairness Testing}

\author{%
  Varsha Ramineni, Hossein A.~Rahmani, Emine Yilmaz, David Barber \\
  Centre of Artificial Intelligence\\
  University College London \\
  \texttt{\{varsha.ramineni.23, hossein.rahmani.22, emine.yilmaz, david.barber\}} \\
  \texttt{@ucl.ac.uk} \\
}

\begin{document}

\maketitle

\begin{abstract}
As AI becomes prevalent in high-risk domains and decision-making, it is essential to test for potential harms and biases. This urgency is reflected by the global emergence of AI regulations that emphasise fairness and adequate testing, with some mandating independent bias audits. However, procuring the necessary data for fairness testing remains a significant challenge. Particularly in industry settings, legal and privacy concerns restrict the collection of demographic data required to assess group disparities, and auditors face practical and cultural challenges in gaining access to data. Further, internal historical datasets are often insufficiently representative to identify real-world biases. This work focuses on evaluating classifier fairness when complete datasets including demographics are inaccessible. We propose leveraging separate overlapping datasets to construct complete synthetic data that includes demographic information and accurately reflects the underlying relationships between protected attributes and model features. We validate the fidelity of the synthetic data by comparing it to real data, and empirically demonstrate that fairness metrics derived from testing on such synthetic data are consistent with those obtained from real data.  This work, therefore, offers a path to overcome real-world data scarcity for fairness testing, enabling independent, model-agnostic evaluation of fairness, and serving as a viable substitute where real data is limited.

\end{abstract}

%

\section{Introduction}
\label{sec:intro}

It is well established that Artificial Intelligence (AI) systems have the potential to perpetuate, amplify, and systemise harmful biases \cite{buolamwini2018gender, caliskan2017semantics}. Therefore, rigorous testing for bias is imperative to mitigate harms, especially given the increasing influence of AI in high-stakes domains such as lending, hiring, and healthcare. Such concerns have fuelled active research in bias detection and mitigation \cite{mehrabi2021survey}, and ensuring the fairness of AI systems has become an urgent policy priority for governments around the world \cite{gov2023pro, aibill}. 
For instance, the EU AI Act imposes strict safety testing on high-risk systems \cite{aiact}, while New York City Local Law 144 mandates independent bias audits for AI used in employment decisions \cite{Groves2024-xo}.

However, procuring the necessary data for fairness testing remains a significant challenge. Influential works in ethics and fairness of machine learning have highlighted the centrality of datasets \cite{jo2020lessons,bao2021s}, emphasising how representative  model testing and evaluation data is crucial \cite{bergman2023representation,shome2024data}. 
To effectively uncover biases, complete datasets that include demographic information and their relationship with model features are essential for controlling the impact of proxy variables. However, having access to such datasets that can reliably be used for evaluating fairness may not always be possible in practice.

As a motivating example, consider a bank that uses an AI system to assess loan applicants based on non-protected variables such as occupation and savings. The bank wants to perform an \emph{internal} audit as to whether its AI system inadvertently discriminates against certain racial groups. For this, the bank requires data concerning protected attributes such as the race of the applicants alongside data of non-protected attributes required by the model to make a loan decision. 

Whilst protected attributes such as race, sex, age etc.~are crucial to assess bias, their collection and use in modelling are heavily restricted under regulations such as GDPR \cite{andrus2021we,veale2017fairer}. Hence, most internal datasets collected by organisations that use AI systems for decision making (such as the bank in our example) do not contain such protected attributes \cite{madaio2022assessing}. Similarly, procuring the necessary data is also a huge complexity for auditors, hindering the effective implementation of algorithmic auditing laws \cite{Groves2024-xo}. In an \emph{external} audit of fairness, the auditing agency often has access only to the black box loan predictions and is not provided any data by the bank since existing regulations often do not allow data holders to release datasets that pause privacy concerns. For this external audit the agency needs a joint distribution of both the attributes needed by the black box loan classifier and the protected attributes. Therefore, the development of curated test sets capable of effectively uncovering biases is essential \cite{madaio2022assessing}. 

Recently, there has been shift away from using limited real test data towards leveraging synthetic data, which has shown promise in a variety of applications ranging from privacy preservation \cite{assefa2020generating} to emulating scenarios for which collecting data is challenging \cite{johnson2016driving}. 

\begin{figure}[t]
\centering
    \includegraphics[width=0.8\textwidth]{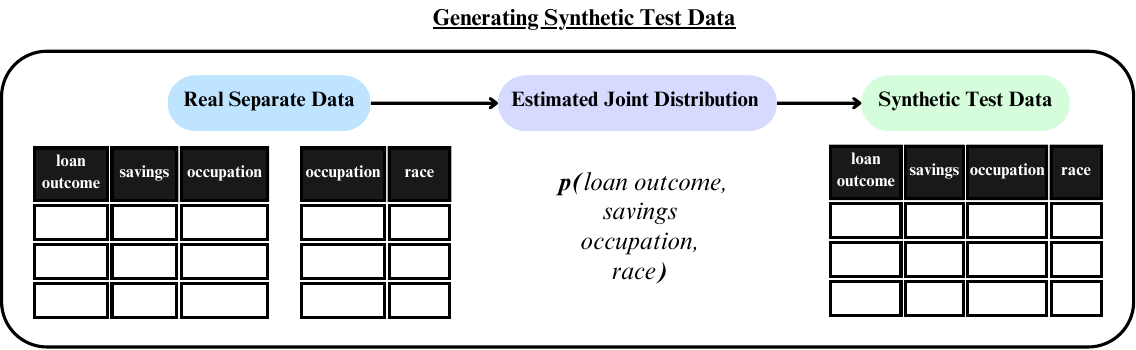}
    \caption{Creation of a synthetic dataset by using two separate datasets and learning their joint distribution. This produces a complete and representative synthetic dataset with essential demographic information necessary for fairness testing.}
    \label{fig:gen-data}
\vspace{-1em}
\end{figure}

Our work focuses on the challenge of evaluating classifier fairness in scenarios where complete data including protected attributes is inaccessible. To overcome this challenge, we propose leveraging separate datasets containing overlapping variables,  which are more accessible in real-world scenarios than complete datasets containing all variables \cite{frogner2019fast}. Specifically, in addition to using an internal dataset that lacks protected attribute information, we propose utilising external data, such as census datasets which provide representative demographic information. For example, the UK Office for National Statistics \cite{ONS2021Census} offers multivariate data from the 2021 Census, providing access to customisable combinations of census variables. Such external data could be utilised when the essential demographic information needed for fairness testing is not directly available.

\begin{figure}[t]
\centering
\includegraphics[width=0.8\textwidth]{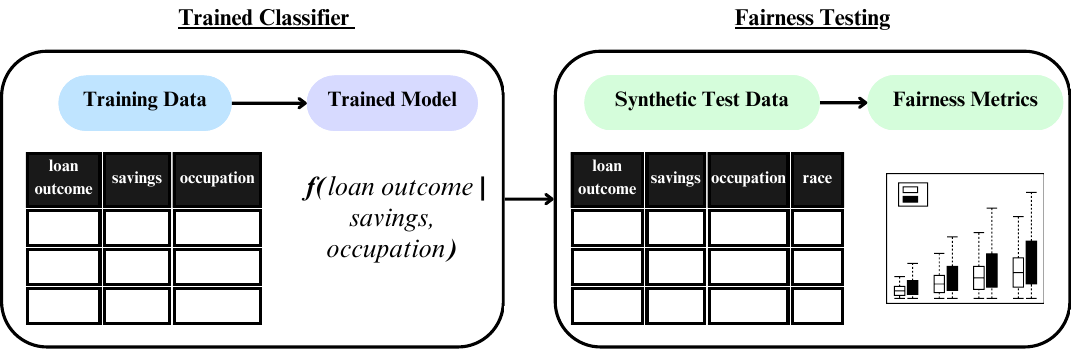}
    \caption{Evaluation of a pre-trained black-box classifier (e.g.~a classifier used by a bank for loan/no-loan decision) on the synthetic data which includes demographics not available during training, enabling the calculation of fairness metrics.
    \label{fig:fair-test}}
\vspace{-1em}
\end{figure}

In our motivating example above, even if the protected attribute `race' is not directly available in the internal dataset, its connection to the features used by the model such as occupation, savings, etc.~can be used to evaluate fairness with respect to race. For instance, the internal dataset used by the bank might include information about $\{loan \, outcome, \; savings, \; occupation\}$. By utilising an external dataset which contains an overlapping variable such as $\{occupation, \; race\}$ that is representative of the population, we can learn the joint distribution of variables from these two datasets, which can then be used to generate synthetic joint test data that contains all the variables, e.g.~$\{loan \, outcome, \; savings, \; occupation, \; race\}$ as shown in \figref{fig:gen-data}. This dataset can then be reliably used for evaluating the fairness of the model, as shown in \figref{fig:fair-test}.

In this work, we conduct experiments on multiple real-world datasets commonly used in fairness research, simulating realistic scenarios involving separated datasets, such as isolated protected attributes and only a single overlapping variable. Our results show that the synthetic test data generated using our proposed approach exhibits high fidelity when compared to real test data. Crucially, we find that fairness metrics derived from testing classifier models on synthetic data closely align with those obtained from real data. These findings suggest that our approach provides a reliable method for fairness evaluation in scenarios where complete datasets are inaccessible, offering a viable alternative for testing in such contexts.

\section{Related Work}
\label{sec:related-work}

\paragraph{Fairness Testing} 
Significant work on fairness evaluation has centered on formalising definitions of fairness \cite{mehrabi2021survey} and emphasising the critical role of data \cite{ bao2021s, jo2020lessons, gebru2021datasheets, paullada2021data}. Recent work has also explored fairness testing in response to regulatory requirements \cite{Groves2024-xo, veale2017fairer} and in the context of industry  \cite{Holstein2019-gt, madaio2022assessing} and software development \cite{chen2021synthetic}. Additionally, there is growing interest in sample-efficient approaches to fairness testing \cite{JiSS20, van2024can}.


\paragraph{Synthetic Data Generation.} Generative models aim to learn the underlying distribution from real data and produce realistic synthetic data. In our work, we focus on tabular data, as it is the most common data type for real-world applications \cite{shwartz2022tabular}. Various models have been developed for tabular data generation, from simple methods like SMOTE \cite{chawla2002smote} to deep learning approaches such as CTGAN \cite{xu2019modeling} and TVAE \cite{xu2019modeling}. Significant previous work has focused on privacy-preserving synthetic data generation, employing marginal-based methods like the MST algorithm \cite{mckenna2021winning}, with work showing that marginal-based algorithms and traditional methods such as mixture models, are more effective at preserving attribute correlations compared to deep learning approaches \cite{richardson2018gans,pereira2024assessment}. Recent innovative advancements also include using large language models \cite{borisovlanguage} and offering customisable tabular data generation \cite{vero2024cuts}. However, these methods typically assume access to full datasets to learn from, limiting their effectiveness in scenarios with restricted data access. 

\paragraph{Synthetic Data for Bias.} Synthetic data for bias has predominately focused on creating fair data for training \cite{xu2018fairgan,van2021decaf}; however, this offers no guarantee of unbiased models \cite{eitan2022fair} and reliable testing methods are therefore crucial. Another approach is to simulate different scenarios to explore the interconnection between biases and their effect on performance and fairness evaluations \cite{baumann2023bias,castelnovo2022clarification}. Recent work highlights the potential of synthetic data for evaluation, showing that, whilst testing on limited real data is unreliable, utilising synthetic test data allows for granular evaluation and testing on distributional shifts \cite{van2024can}. Emerging work also looks at most effective synthetic data generation techniques for training and evaluating machine learning models and the implications of model fairness \cite{pereira2024assessment}. 



\section{Methodology}
\label{sec:methodology}
Returning to our motivating example of a loan classifier, our assumption is that the classifier uses only non-protected attributes, such as savings $X$, and occupation $O$ in order to form a loan prediction $\hat{Y}$; in this case, the loan decision is some function of the non-protected attributes, e.g.~$\hat{Y} = f(X, O)$. However, we would like to assess whether this prediction is fair against a protected attribute $A$ such as race. There are various statistical definitions of group fairness in classification, typically conditioned on protected attributes along which fairness should be ensured. We use the following notation: Let $Y \in \{+,-\}$ represent the true outcome, $\hat{Y} \in \{+,-\}$ the predicted outcome, and $A \in \{privileged, unprivileged\}$ the protected attribute. Here, `$+$' denotes a positive classification outcome (e.g., loan approval), while `$-$' denotes a negative outcome (e.g., loan rejection). For instance, the fairness metric \textbf{Equal Opportunity Difference} (EOD) is given by:
\begin{equation}
    EOD = P(\hat{Y} = + \mid Y = +, A = unprivileged) - P(\hat{Y} = + \mid Y = +, A = privileged) 
    \label{Eq:EOD}
\end{equation}
To calculate this, one necessary term is $ P(\hat{Y}=+ \mid Y=+, A),$ where 
\begin{equation}
 P(Y=+, A) = \sum_{O, X} P(Y=+, O, X, A).   
\end{equation}
This requires a model of the joint distribution (as shown in \figref{fig:gen-data}), which can then be used to test the fairness of a pre-trained black-box classifier, as illustrated in \figref{fig:fair-test}. In the following section, we explain how to construct joint distributions from a collection of overlapping marginal distributions.

\subsection{Learning a Joint Distribution}
\label{sec:learning-a-joint-distribution}
Consider a fairness testing scenario that requires access to the distribution $p(loan \, outcome, savings, occupation, race)$. Most real-world datasets, such as provided by publicly available census data, often only provide sets of marginal distributions \cite{frogner2019fast}. Suppose we have two separate datasets with empirical distributions $\hat{p}(loan \, outcome, savings, occupation)$ and $\hat{p}(occupation, race)$, where $occupation$ is the overlapping variable. Our goal is to estimate the joint distribution $ p(loan \, outcome, savings, occupation, race)$. Theoretically, this problem is ill-posed and therefore requires additional assumptions.  

Using marginal data observations and a structural independence assumption, the joint distribution can be estimated using maximum likelihood estimation. We consider below three simple structural independence assumptions, illustrated by graphical models, to fit a joint distribution on four variables $p(x_1,x_2,x_3,x_4)$, given two empirical marginal distributions $\hat{p}(x_1,x_2,x_3)$ and $\hat{p}(x_3,x_4)$. The estimated joint distribution is then used as a generative model to create synthetic data points through sampling \cite{vora2021recovery}. Note that we assume marginal consistency i.e.~that all marginal distributions considered originate from a common underlying joint distribution.

\subsubsection{Independence Given Overlap}

\begin{minipage}{0.45\linewidth}
\begin{tikzpicture}[
    vertex/.style={draw=black,fill=white, ellipse, minimum height=0.1cm, minimum width=0.1cm},
    singlevertex/.style={circle,draw=black,fill=white},
    node distance=2cm,
    on grid,
    >=latex
]
    \node[singlevertex] (A) {$X_3$};
    \node[vertex, left=1.5cm and 2.5cm of A] (B) {$X_1, X_2$};
    \node[singlevertex, right=1.5cm and 2cm of A] (D) {$X_4$};
    \draw[->]
        (A) edge (B)
        (A) edge (D);
\end{tikzpicture}
\end{minipage}
\begin{minipage}{0.55\linewidth}
\begin{equation}
p(x_1, x_2, x_3, x_4) =  p(x_3) \cdot  \hat{p}(x_1, x_2 | x_3) \cdot \hat{p}(x_4 | x_3)
\label{method1}
\end{equation}
\end{minipage}

We model the joint distribution of $x_1, x_2, x_3$, and $x_4$ by treating the association between $(x_1, x_2)$ and $x_4$ as the product of their conditional distributions given $x_3$. To estimate $p(x_3)$, we take the average of the proportions from both marginal datasets and use this to sample $x_3$ (see \appref{appendix:independence-given-overlap} for proof of optimality). To sample from this model, we first sample from $p(x_3)$ and then draw conditional samples for $(x_1, x_2)$ and $x_3$ from the marginal datasets. Note that if the marginals are consistent, namely $\sum_{x_1, x_2}\hat{p}(x_1,x_2, x_3) = \sum_{x_4}\hat{p}(x_3, x_4) \equiv \hat{p}(x_3)$, then we simply set $p(x_3)=\hat{p}(x_3)$.

\subsubsection{Marginal Preservation}

\begin{minipage}{0.5\linewidth}
\centering
\begin{tikzpicture}
  [vertex/.style={circle,draw=black,fill=white},
   node distance=2cm,
   on grid,
   >=latex
  ]
    \node[vertex] (A) {$X_1$};
    \node[vertex, right=1.5cm of A] (B) {$X_2$};
    \node[vertex, right=1.5cm of B] (C) {$X_3$};
    \node[vertex, right=1.5cm of C] (D) {$X_4$};
  \draw[->]
    (C) edge (D);
    \draw[-]
    (A) -- (B)
    (C) -- (B);
\end{tikzpicture}
\end{minipage}
\begin{minipage}{0.5\linewidth}
\begin{equation}
p(x_1, x_2, x_3, x_4) = \hat{p}(x_1, x_2, x_3) \cdot  \hat{p}(x_4 | x_3) 
\label{method2}
\end{equation}
\end{minipage}

We directly use the proportions from the first marginal dataset to model the joint distribution of $x_1$, $x_2$ and $x_3$. A sample is then obtained by sampling from the marginal $\hat{p}(x_1,x_2, x_3)$ and then from the conditional marginal $\hat{p}(x_4|x_3)$. Alternatively, we could preserve the second marginal by modelling the distribution as $p(x_1, x_2, x_3, x_4) = \hat{p}(x_1, x_2 | x_3) \cdot \hat{p}(x_3, x_4) $.

\subsubsection{Latent Naïve Bayes} 

\begin{minipage}{0.5\linewidth}
\centering
\begin{tikzpicture}%
  [vertex/.style={circle,draw=black,fill=white},
   node distance=2cm,
   on grid,
   >=latex
  ]
  \node[vertex] (Z) {$Z$};
  \node[vertex,below left=1.25cm and 2cm of Z] (A) {$X_1$};
  \node[vertex,below left=1.25cm and 0.7cm of Z] (B) {$X_2$};
  \node[vertex,below right=1.25cm and 0.7cm of Z] (C) {$X_3$};
  \node[vertex,below right=1.25cm and 2cm of Z] (D) {$X_4$};
  \draw[->]
    (Z) edge (A)
    (Z) edge (B)
    (Z) edge (C)
    (Z) edge (D);
\end{tikzpicture}
\end{minipage}
\begin{minipage}{0.5\linewidth}
\begin{equation}
    p(x_1, x_2, x_3 ,x_4) = \sum_z p(z) \prod_{i=1}^4  p(x_i | z)
    \label{eq:method3}
\end{equation}
\end{minipage}

We employ a latent variable model based on the Naïve Bayes assumption by introducing a latent variable $z$, which assumes that $x_1$, $x_2$, $x_3$, and $x_4$ are conditionally independent given $z$. We use the Expectation-Maximization (EM) algorithm \cite{dempster1977maximum} to train the model (see \appref{appendix:latent-naive-bayes} for details). 

\subsubsection{Extension to More Complex Scenarios}

We can extend the Latent Naïve Bayes method to include more variables by adding the term \( p(x_k \mid z) \) for any new variable \( x_k \). Similarly, other methods can be adapted to handle additional variables. For instance, if the second marginal distribution is \( \hat{p}(x_3, x_4, x_5, x_6) \), we adjust the conditional distribution from \( \hat{p}(x_4 \mid x_3) \) to \( \hat{p}(x_4, x_5, x_6 \mid x_3) \). When multiple variables overlap between datasets, such as in the empirical distributions \( \hat{p}(x_1, x_2, x_3, x_4) \) and \( \hat{p}(x_3, x_4, x_5) \) where \( (x_3, x_4) \) are overlapping, we extend the methods to preserve the joint structure. For the \textit{Independence Given Overlap} method, we use:
$
p(x_1, x_2, x_3, x_4, x_5) = p(x_3, x_4) \cdot \hat{p}(x_1, x_2 \mid x_3, x_4) \cdot \hat{p}(x_5 \mid x_3, x_4).
$
For the \textit{Marginal Preservation} method, we use:
$
p(x_1, x_2, x_3, x_4, x_5) = \hat{p}(x_1, x_2, x_3, x_4) \cdot \hat{p}(x_5 \mid x_3, x_4). 
$ 

In this work, we focus on estimating the joint distribution from two datasets that overlap in a single variable. Real-world datasets may exhibit more complex structures involving multiple datasets. While Latent Naïve Bayes offers a straightforward extension to multiple datasets, there could be alternative approaches such as using Junction Trees \cite{barber2021}. Such work is left for future research, with this study serving as a preliminary exploration of our proposed approach.

\section{Experimental Setup}

\begin{figure*}[t]
\centering
\includegraphics[width=0.8\textwidth]{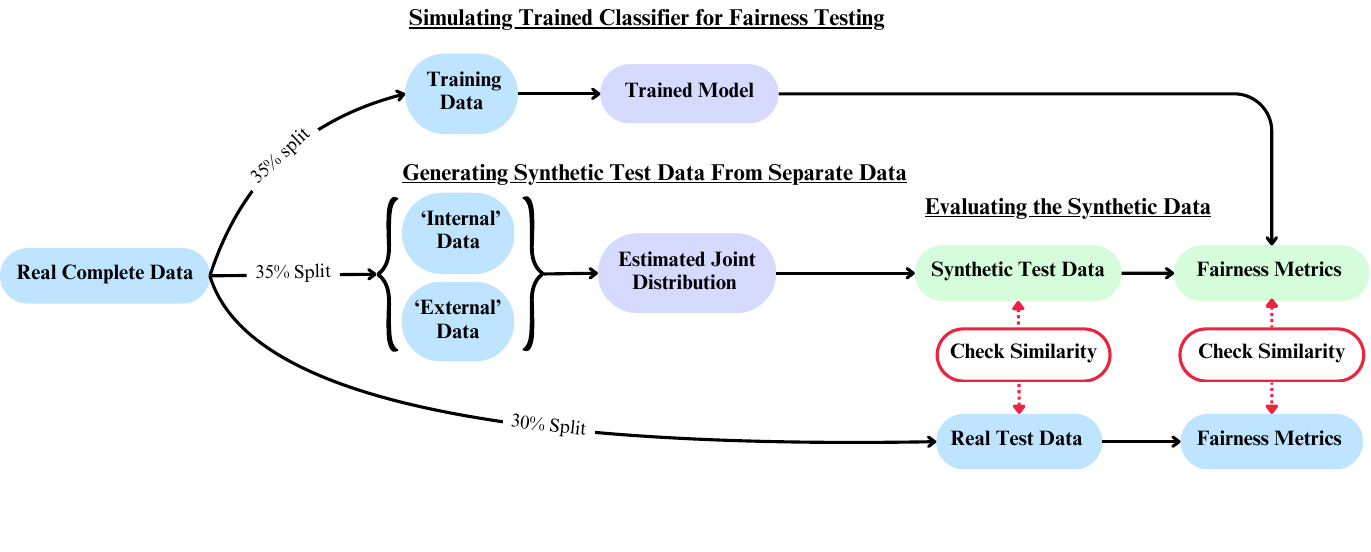} 
\vspace{-1em}
\caption{Experimental Setup }
\label{fig:exp-setup}
\vspace{-1em}
\end{figure*}

We aim to generate synthetic datasets and evaluate their quality based on two criteria: 1) how well they can approximate a real ground truth dataset, and 2) how accurately they can estimate the fairness of a black-box classifier in situations where complete data, including protected attributes is inaccessible. We assume that, as in our example, we have access to two separate datasets, for example one containing $\{loan \, outcome, savings, occupation\}$ and another containing $\{occupation, race\}$, used to estimate a joint distribution and generate a synthetic test dataset including all attributes. In this setup, one dataset includes the protected attribute, while the other contains model input features, with an overlapping variable between the two datasets. 

\subsection{Datasets}
We conduct our experiments using three real-world datasets: \textbf{Adult} \cite{adult_2}, \textbf{COMPAS} \cite{compas}, and \textbf{German Credit} \cite{dheeru2017uci}, detailed in \tabref{tab:data}, which are commonly used in the fairness literature. For all three datasets we follow the literature by removing instances with null values, and map all continuous variables into categorical variables (see \appref{appendix:datasets} for details) \cite{le2022survey}.  These datasets represent complete real data with protected attributes. Our goal is to approximate such data using our synthetic data generation approach.

\begin{table}
  \caption{Overview of real world datasets used in experiments}
  \label{tab:data}
  \centering
  \begin{tabular}{lllll}
    \toprule
    \textbf{Name}     & \textbf{\# Instances}     & \textbf{\# Attributes}  & \textbf{Label} & \textbf{Protected Attributes} \\
    \midrule
    Adult \cite{adult_2} & 45,222  & 13  & Income & Sex (67.5\% male, 32.5\% female) \\
    &&&&  Race (86\% white,14\% non-white) \\
    COMPAS \cite{compas} & 5278  & 9  &  Recidivism  & Sex (80.5\% male, 19.5\% female) \\
    &&&& Race (60.2\% white, 39.8\% black) \\
    German \cite{dheeru2017uci}    &  1000       & 22 & Credit Risk & Age (81\% > 25, 19\% $\leq$ 25) \\
    &&&& Sex (69\% male, 31\% female)\\
    \bottomrule
  \end{tabular}
\vspace{-1em}
\end{table}

\begin{table}[t]
\caption{Separation of complete real datasets, with each row illustrating how attributes are categorised into `external' and `internal' datasets. The `external' dataset shown includes protected attributes, while the `internal' dataset comprises the remaining attributes. Protected attributes are shown in bold, and overlapping variables shared between the two datasets are shown in italics. }
\label{tab:separate-data}
\begin{center}
\begin{tabular}{ll}
\toprule
    \textbf{Dataset} & \textbf{Attributes in `External' Dataset (overlapping variable in \textit{italics})}  \\
    \midrule
    Adult  & \textbf{\textit{relationship}, age, sex, race, marital-status, native-country} \\
    & \textbf{\textit{marital-status}, age, sex, race, marital-status, native-country} \\
    COMPAS   & \textit{score}, \textbf{sex, age, race} \\
       & \textit{violent score}, \textbf{sex, age, race} \\
    German Credit  &  \textit{property}, \textbf{sex, marital-status, age, foreign-worker} \\
    &  \textit{housing}, \textbf{sex, marital-status, age, foreign-worker} \\
\bottomrule
\end{tabular}
\end{center}
\vspace{-1em}
\end{table}

\subsection{Simulating Data Scenarios}
Our experimental setup is visualised in \figref{fig:exp-setup}. To assess our approach, we simulate having a known ground truth dataset to compare our generated synthetic data against.

\paragraph{Real Test Data.} 
Starting with a complete real dataset, we reserve a hold-out real test set $D_{\textit{test}}$ (30\% of the complete real dataset) that includes all relevant attributes. This is the dataset that we would like to approximate using the synthetic data we generate and we use this to assess our approach.

\paragraph{Separated Data.}
We wish to simulate the scenario where we don't have access to complete data but only have two separate datasets as illustrated in \figref{fig:gen-data}. We therefore separate the remaining complete real data by column into two overlapping datasets. We consider separations where protected attributes are isolated from other variables, and where there is one variable overlapping between datasets. We refer to these separate datasets as `internal' and the `external', where the `external' data includes protected attributes not available in the `internal' data. Such separation simulates only having access to protected attributes separately, such as in publicly available census data, and assumes limited overlap of attributes. 

\Tabref{tab:separate-data} demonstrates the separation of our three complete real-world datasets. Notably, the `external' datasets includes data commonly found as census variables. As illustrated in \figref{fig:gen-data}, we use the two separate datasets to estimate the joint distribution of all attributes and generate synthetic test data $D_{\textit{synth}}$. We also wish to simulate having a trained classifier that we wish to test for fairness, as shown in \figref{fig:fair-test}. This is done by training classifier models on one of the real separate datasets, the `internal' dataset, which does not include protected attributes. The classifier models will then be tested on both synthetic and real test data.

\subsection{Baselines}
\label{sec:baselines}
To our knowledge, no prior work for fairness testing has tackled the challenge of creating synthetic data from separate datasets that accurately capture the relationship between demographic and model features. We compare our approach with common methods for tabular synthetic data generation. The \textbf{Independent Model} assumes independence between any two variables to estimate the joint distributions \cite{mckenna2022aim}. \textbf{Conditional Tabular GAN (CTGAN)} \cite{xu2019modeling} is a state-of-the-art method that learns from the full dataset, unlike our method, which works with separate datasets. Although CTGAN has an advantage due to its access to complete data, we include its performance using default hyperparameters for comparison.

\section{Evaluating the Quality of Synthetic Datasets}
We use two criteria to evaluate the quality of our synthetic datasets: 1) How does the synthetic data compare with real data? and 2) How does the fairness metrics computed on the synthetic data compare with real data? 

We present results for eighteen synthetic test datasets which were generated using the three joint distribution estimation methods, applied to six different pairs of separated data across real world datasets: Adult, COMPAS and German Credit. \Tabref{tab:separate-data} shows an overview of how our datasets have been separated and which overlapping attributes have been used. In addition to the synthetic datasets generated using our proposed approach, we also generate synthetic datasets using the two baseline approaches and compare the quality of our synthetic datasets with the quality of synthetic datasets generated using the baseline methods. 

\begin{table}[t]
\caption{Fidelity metrics for synthetic datasets of the Adult dataset, generated from separate data ( `relationship' overlapping) with different joint estimation methods. Metrics include total variation distance complement (1-TVD), contingency similarity (CS), discriminator measure (DM), and KL divergence of $p(A,Y)$ in synthetic vs real data (where Y is the outcome label and A is a protected attribute such as race and sex). Baseline methods include CTGAN and Indep.}
\centering
\begin{tabular}{lcccccccccc}
\toprule
\multirow{2}{*}{Method} & \multicolumn{3}{c}{Overall Fidelity} & \multicolumn{2}{c}{Joint Distribution for (A,Y)}  \\ 
\cmidrule{2-4} \cmidrule{5-8}
& \textbf{1-TVD} $\uparrow$ & \textbf{CS} $\uparrow$ & \textbf{DM} $\downarrow$ & \textbf{KL (Race)} $\downarrow$ & \textbf{KL (Sex)} $\downarrow$ \\
\midrule
Indep-Overlap (Relationship) & 0.993 & 0.983 & 0.588 & 0.002 & 0.001  &   \\
Marginal (Relationship) &  0.993 & 0.983 & 0.588 & 0.002 & 0.001  \\
Latent (Relationship) & 0.986 & 0.968 & 0.658 & 0.002 & 0.002  \\
CTGAN & 0.935 & 0.938 & 0.656 & 0.132 & 0.048 \\
Independent & 0.935 & 0.895 & 0.808 & 0.005 & 0.026  \\
\bottomrule
\end{tabular}
\label{tab:fidelity_main1}
\end{table}

\begin{figure}[t]
    \centering
    
    \begin{minipage}[c]{\textwidth}
        \centering
        \includegraphics[width=\textwidth]{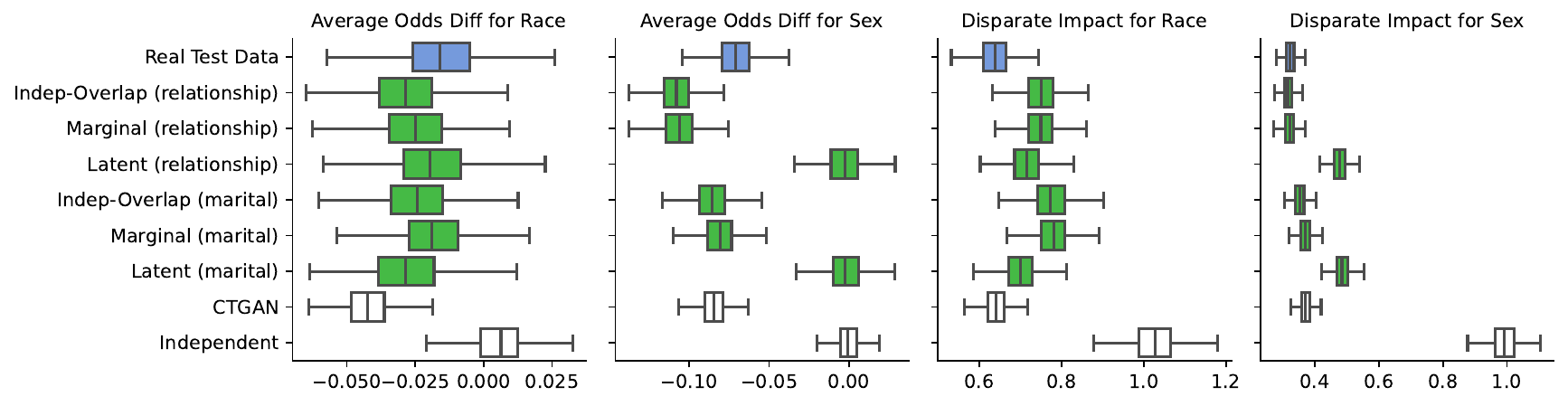}
        \text{(a) Adult Data}
        \label{fig:first}
    \end{minipage}
        
    \begin{minipage}[c]{\textwidth}
        \centering
        \includegraphics[width=\textwidth]{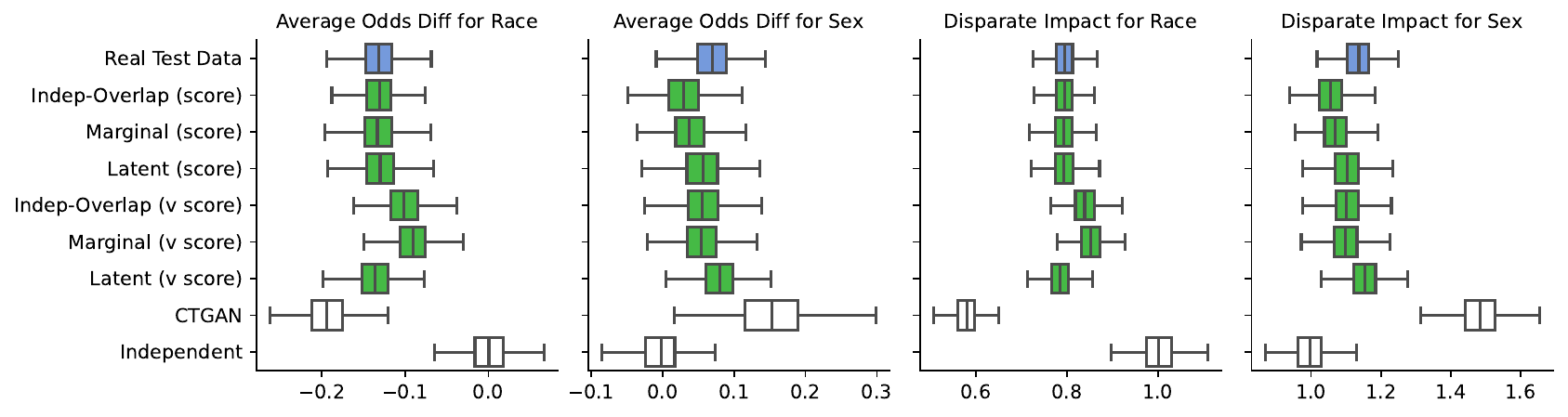}
        \text{(b) COMPAS Data}
        \label{fig:second}
    \end{minipage}
        
    \begin{minipage}[c]{\textwidth}
        \centering
        \includegraphics[width=\textwidth]{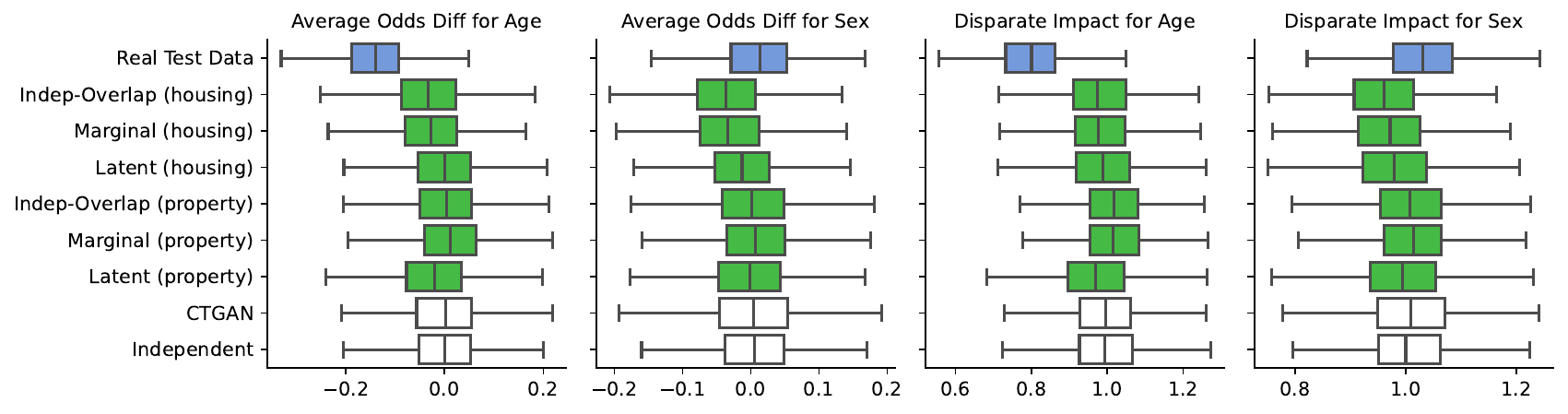}
        \text{(c) German Credit Data}
        \label{fig:third}
    \end{minipage}
    \caption{Box-plots of fairness metrics for a Decision Tree Classifier across synthetic datasets. Each subplot represents a specific fairness metric for a protected attribute, showing the distribution of metrics from bootstrap samples. The top box-plot in each subplot displays the distribution of the metric from testing the classifier on real test data (blue), the middle boxplots (green) are for synthetic data generated using our approach differentiated by data separation methods (overlapping variables in brackets) and joint estimation methods, and the bottom box plots are for baseline methods (white).}
    \label{fig:fair_met}
\vspace{-1em}
\end{figure}

\subsection{Overall Fidelity of Synthetic Data Compared to Real Data}

Fidelity evaluates how close the distribution of the synthetic data is to that of the real data with metrics often estimating the difference between marginal distributions \cite{patki2016synthetic,platzer2021holdout,tao2021benchmarking}. To evaluate the fidelity of our synthetic datasets, we focus on the following metrics:

\begin{itemize}
    \item \textbf{Total Variation Distance (TVD)}: Measures the difference between the empirical distributions of a variable in the synthetic data and the real data, defined as half the $L_1$ distance. We use the TVD Complement score, $1 - \text{TVD}$, where higher scores close to one indicate better quality synthetic data (averaged across variables) \cite{patki2016synthetic}. 
    
    \item \textbf{Contingency Similarity (CS)}: Assesses the similarity between normalised contingency tables of two variables, one from the real data and one from the synthetic data. This metric is calculated by first normalising the contingency tables to show the proportion of each category combination, then computing the TVD between these tables. The complement, $1 - \text{TVD}$, is used so that higher values close to one reflect greater similarity (averaged across varibales) \cite{patki2016synthetic}. 

    \item \textbf{Cramér's V Correlation}: Quantifies the strength of association between two categorical variables based on the Chi-square statistic \cite{cramér1946mathematical}. We calculate the difference in Cramér's V correlation between the synthetic and real data for each pair of variables. 

    \item \textbf{Discriminator Measure (DM)}: Evaluates whether the synthetic data can be distinguished from the real data. We train a Random Forest Classifier on a balanced dataset, with synthetic data labeled as $1$ and real data labeled as $0$. The classifier's average accuracy on a test set is reported across five trials with different random seeds \cite{borisovlanguage}.
\end{itemize}

The eighteen synthetic datasets generated using our approach demonstrate high fidelity to the real test data, with an example shown in \tabref{tab:fidelity_main1} for the Adult dataset. The average $1-\text{TVD}$ values across synthetic data for the Adult, COMPAS, and German datasets are 0.991, 0.978, and 0.966 respectively, while the average CS values are 0.978, 0.953, and 0.926. These results demonstrate the effectiveness of our approach in generating data that closely mirrors the proportions of the real test dataset. The results also show competitive or superior performance compared to the CTGAN baseline method, which generates synthetic data from complete data rather than separate data. The DM scores reveal moderate accuracy in distinguishing synthetic from real data. Across the eighteen synthetic datasets there is on average a 12.9\% reduction in discriminator performance compared to the Independent Baseline and an 8.2\% reduction compared to the CTGAN Baseline, suggesting that the synthetic test data is more challenging to differentiate from real data. Additionally, the difference in Cramér's V correlations between synthetic and real datasets suggests that the attribute correlations in our synthetic data closely match those in the real data, showing greater similarity than baseline methods. See \appref{appendix:fidelity} for correlation figures and full fidelity results.


\subsection{Protected Attribute and Outcome Relationship in Synthetic Data Compared to Real Data}
As illustrated in \secref{sec:methodology}, understanding the relationship between the protected attribute \( A \) and the outcome label $Y$ is essential for assessing group disparities. When $A$ and $Y$ are located in separate datasets, such as the simple case in our loan example, 
it is crucial that the relationship between these variables $(A,Y)$ is accurately reconstructed in the synthetic datasets. We therefore measure the Kullback-Leibler (KL) divergence, $D_{\text{KL}}(p_{\textit{synth}}(A,Y) \parallel p_{\textit{real}}(A,Y))$, between the joint distributions $p(A, Y)$ of synthetic and real data. KL divergence values close to zero indicate that the joint distribution of protected attribute and outcome label in the synthetic data is similar to the distribution in the real data

\tabref{tab:fidelity_main1} presents the divergence for the Adult dataset, focusing on synthetic data generated from separate data which had `relationship' as the overlapping variable. Across all separations and joint distribution estimation methods for Adult Data, the average KL divergence is 0.002 for Race and 0.001 for Sex. Despite generating synthetic data from separate datasets with only one overlapping variable, the joint distribution of protected attributes and outcome values is accurately reconstructed, as evidenced by the low KL divergence values. In comparison, CTGAN shows higher KL divergence values of 0.132 for Race and 0.048 for Sex. Similar patterns are observed in across the other datasets, with detailed results provided in \appref{appendix:kl}.

\subsection{Fairness Metrics from Synthetic Data Compared to  Real Data}
\label{sec:fairness-metrics-on-synthetic-vs-real-data}

We next compare how fairness metrics computed on synthetic test datasets compare with those from real test datasets. Using the notation from \secref{sec:methodology}, we focus on the Equal Opportunity Difference (EOD) (Equation~\ref{Eq:EOD}) and two other common metrics: Disparate Impact (DI) and Average Odds Difference (AOD) \cite{mehrabi2021survey}. The \textbf{Disparate Impact (DI)} metric compares the ratio of positive (favorable) outcomes between the unprivileged and the privileged groups and can be computed as: 
\begin{equation}
\text{DI} = \frac{p(\hat{Y} = + | A = unprivileged)}{p(\hat{Y} = + | A = privileged)}
\end{equation}

The \textbf{Average Odds Difference (AOD)} metric measures the disparity between the false positive rate and true positive rate for the unprivileged and privileged groups and can be written as follows:
\begin{align}
  \text{AOD} = &\frac{1}{2} \Bigg[ p(\hat{Y}=+ \mid Y=-, A=unprivileged) - p(\hat{Y}=+ \mid Y=-, A=privileged) \nonumber \\
  & + p(\hat{Y}=+ \mid Y=+, A=unprivileged) - p(\hat{Y}=+ \mid Y=+, A=privileged) \Bigg]
\end{align}

\figref{fig:fair_met} compares fairness metrics between synthetic and real test datasets for a Decision Tree classifier. For each dataset, we generate 1,000 bootstrap samples of the same size as the real test data to compute fairness metrics. Box-plots for DI and AOD illustrate the distribution of these metrics. EOD, which trends similarly to AOD, is omitted from the figure but included in \appref{appendix:fairness-metrics} with detailed results on the absolute differences between bootstrap means of fairness metrics from synthetic and real data. The results show that the fairness metrics from our synthetic test data closely match those from real data, outperforming baseline methods on nearly all metrics and protected attributes, except for DI for race in the Adult dataset. Notably, the synthetic data for the COMPAS dataset performs best, with absolute differences of 0.000 in bootstrap means for AOD and DI values for race, achieved using the `Marginal' joint estimation method on separate data with the `violent score' variable overlapping. For the Adult dataset, we also see small absolute differences in bootstrap means, with values as low as 0.002 for DI related to sex, 0.003 for AOD related to race, and 0.010 for AOD related to sex. For the German dataset, we see similar results, showing small absolute differences of 0.005 for AOD and 0.015 for DI related to sex. Despite larger differences shown in fairness metrics for age, the synthetic data still outperforms baseline methods.

\section{Conclusion and Future Work}

In this study, we tackled the challenge of evaluating classifier fairness when complete datasets, including protected attributes, are inaccessible. We proposed an approach that utilises separate overlapping datasets to estimate a joint distribution and generate complete synthetic test data which includes demographic information and accurately captures the relationships between demographics and model features essential for fairness testing. Our empirical analysis demonstrated that the fairness metrics derived from this synthetic test data closely match those obtained from real data. Our results further show that even with the assumption of only a single overlapping variable between separate datasets, and simple joint distribution estimation methods, the synthetic data can closely mirror real data outcomes and exhibit high fidelity. 

This work demonstrates a promising approach for fairness testing by leveraging marginally overlapping datasets to curate effective test datasets. However, we simulated separate datasets and data scenarios, future research could explore incorporating real public data and more complex data scenarios to validate the results obtained. We also employed three joint estimation methods using structural assumptions. Future research could instead explore all feasible joint distributions that meet the constraints of the available marginal distributions, and thus work towards defining  bounds within which the true fairness metrics are likely to fall.

\begin{ack}

The authors declare no competing interests related to this paper. This research was supported by the UKRI Engineering and Physical Sciences Research Council (EPSRC) [grant numbers EP/S021566/1 and EP/P024289/1].


\end{ack}

\bibliography{references}

\clearpage

\appendix

\section*{Appendix }

Appendix related to paper: \textit{Beyond Internal Data:
Constructing Complete Datasets for Fairness Testing} for the Algorithmic Fairness through the Lens of Metrics and Evaluation (AFME) at NeurIPS 2024

The appendix is structured as follows:
\paragraph{Appendix \sect{appendix:learning-joint-distribution}} provides technical details on the joint distribution estimation methods, Marginal Preservation and Latent Naïve Bayes, as outlined in the main text.

\paragraph{Appendix \sect{appendix:datasets}} describes each dataset used in the experiments, with tables specifying the variables and categories present after preprocessing.

\paragraph{Appendix \sect{appendix:quality}} presents detailed  results of the metrics used to assess the quality of the generated synthetic data.

\section{Technical Details for Joint Distribution Estimation Methods}
\label{appendix:learning-joint-distribution}

\subsection{Proof for Optimality for Independence Given Overlap Method}
\label{appendix:independence-given-overlap}

To find the optimal \( p(x_3) \), we start by minimising the total Kullback-Leibler (KL) divergence:

\begin{equation}
  \mathcal{L}(p) = \KL \big(\hat{p}(x_1, x_2, x_3) \parallel p(x_1, x_2, x_3)\big) + \KL \big(\hat{p}(x_3, x_4) \parallel p(x_3, x_4)\big).  
\end{equation}

Let \( \hat{p}_1(x_3) \) and \( \hat{p}_2(x_3) \) be the empirical marginals from the first and second datasets, respectively:

\begin{equation}
\hat{p}_1(x_3) = \sum_{x_1, x_2} \hat{p}(x_1, x_2, x_3), \quad \hat{p}_2(x_3) = \sum_{x_4} \hat{p}(x_3, x_4).
\end{equation}

From our joint distribution assumption
$
p(x_1, x_2, x_3, x_4) = p(x_3) \cdot \hat{p}(x_1, x_2 \mid x_3) \cdot \hat{p}(x_4 \mid x_3)
$, we obtain marginals $
p(x_1, x_2, x_3) = p(x_3) \cdot \hat{p}(x_1, x_2 \mid x_3),
$ and
$
p(x_3, x_4) = p(x_3) \cdot \hat{p}(x_4 \mid x_3).
$

To minimise the KL divergence with respect to \( p(x_3) \), we rewrite \( \mathcal{L}(p) \) focusing on the marginal \( p(x_3) \):

\begin{align}
   \mathcal{L}(p) &= \sum_{x_1, x_2, x_3} \hat{p}(x_1, x_2, x_3) \left[\log \frac{\hat{p_1}(x_3) \hat{p}(x_1, x_2| x_3)}{p(x_3)\hat{p}(x_1, x_2| x_3)} )\right] + \sum_{x_3, x_4} \hat{p}(x_3, x_4) \left[\log \frac{\hat{p_1}(x_3) \hat{p}(x_4| x_3)}{p(x_3)\hat{p}(x_4 | x_3)} )\right] \nonumber \\
    &= - \sum_{x_3} \sum_{x_1, x_2}  \hat{p}(x_1, x_2, x_3) \log p(x_3) - \sum_{x_3} \sum_{x_4}  \hat{p}(x_3, x_4) \log p(x_3) \nonumber \\
    &= - \sum_{x_3} \left(\hat{p}_1(x_3) + \hat{p}_2(x_3)\right) \log p(x_3)
\end{align}

We find the optimal \( p(x_3) \), which minimises $\mathcal{L}(p)$ subject to $\sum_{x_3} p(x_3) = 1$ to ensure that \( p(x_3) \) is a valid probability distribution.

\begin{align}
p(x_3) \propto \hat{p}_1(x_3) + \hat{p}_2(x_3).
\end{align}

To normalise \( p(x_3) \), we set:

\begin{align}
p(x_3) &= \frac{\hat{p}_1(x_3) + \hat{p}_2(x_3)}{\sum_{x_3'} \left(\hat{p}_1(x_3') + \hat{p}_2(x_3')\right)} \nonumber \\
 &= \frac{\hat{p}_1(x_3) + \hat{p}_2(x_3)}{2}
\end{align}

ensuring that \( p(x_3) \) is a valid probability distribution. 

Therefore the optimal $p(x_3)$ is the average of the empirical marginals from both datasets.

\subsection{Details for Expectation Maximisation Algorithm for Latent Naïve Bayes Method}
\label{appendix:latent-naive-bayes}

We assume categorical variables $X_1, X_2, X_3, X_4$ with dom$(X_i) = \{1, 2, ..., M_i\}$ where $M_i \in \mathbb{N}, M_i > 1$ for $i = \{1, 2, 3, 4\}$.  We want to sample from the full joint distribution $p(X_1, X_2, X_3, X_4)$. However, our observations are of the form $\mathbf{D_1}$, and $\mathbf{D_2}$, where $x_3$ and $x'_3$ are both observations of the same variable $X_3$.

\begin{align}
\mathbf{D_1} &= \{({x_{1}}^n, {x_{2}}^n, {x_{3}}^n)\}_{n=1}^{N_1} \\
\mathbf{D_2} &= \{({x'_{3}}^n, {x_{4}}^n)\}_{n=1}^{N_2} 
\end{align}

To model the complex dependencies between the variables and to simplify the model, we intentionally introduce latent variable $Z$, and the following probabilistic graphical model, where dom$(Z) = \{1, 2, . . .,  K\}, K \in \mathbb{N}, K > 1$.

\begin{center}
\begin{tikzpicture}%
  [vertex/.style={circle,draw=black,fill=white},
   node distance=2cm,
   on grid,
   >=latex
  ]
  \node[vertex] (Z) {$Z$};
  \node[vertex,below left=1.25cm and 2cm of Z] (A) {$X_1$};
  \node[vertex,below left=1.25cm and 0.7cm of Z] (B) {$X_2$};
  \node[vertex,below right=1.25cm and 0.7cm of Z] (C) {$X_3$};
  \node[vertex,below right=1.25cm and 2cm of Z] (D) {$X_4$};
  \draw[->]
    (Z) edge (A)
    (Z) edge (B)
    (Z) edge (C)
    (Z) edge (D);
\end{tikzpicture}
\end{center}

By treating $Z$ as a missing variable, mixture models can be trained using the EM algorithm.

The model defines the generative process for each data item $n$ as follows:

\begin{enumerate}
    \item Sample $Z$ from $p(Z = k) = \pi_k$, where $k = 1, 2, \ldots, K$, $\pi_k \geq 0$, and $\sum_{k=1}^K \pi_k = 1$.
    
    \item Given $Z = k$, the conditional distribution of $X_i$ for $i = 1, 2, 3, 4$ is:
    \begin{align}
    p(X_i = m \mid Z = k) = p_i(m \mid k)
    \end{align}
    where $m = 1, 2, \ldots, M_i$, $p_i(m \mid k) \geq 0$, and $\sum_{m=1}^{M_i} p_i(m \mid k) = 1$.
\end{enumerate}

We aim to learn the parameters $\boldsymbol{\theta} = (\theta_1, \theta_2, \theta_3, \theta_4, \theta_Z)$, where:
\begin{align*}
    \theta_i &= \{ p_i(m \mid k) : m = 1, \ldots, M_i, \; k = 1, \ldots, K \} \quad \text{for } i = 1, 2, 3, 4 \\
    \theta_Z &= (\pi_1, \ldots, \pi_K)
\end{align*}

By learning $\boldsymbol{\theta}$, we can model the joint distribution:

\begin{align}
p_{\boldsymbol{\theta}}(X_1, X_2, X_3) = \sum_{Z=1}^{K} p_{\theta_Z}(Z) \prod_{i=1}^{4} p_{\theta_i}(X_i \mid Z)
\end{align}
    
\subsubsection{Model Distributions}

For dataset $\mathbf{D_1}$, the joint distribution is:

\begin{align}
p_{\boldsymbol{\theta}}(\mathbf{D_1}, \mathbf{z}) = \prod_{n=1}^{N_1} p_1(x_1^n \mid z^n) \cdot p_2(x_2^n \mid z^n) \cdot p_3(x_3^n \mid z^n) \cdot \pi_{z^n}  
\end{align}

Marginalising over the latent variables gives the marginal log likelihood:

\begin{align}
\log p_{\boldsymbol{\theta}}(\mathbf{D_1}) = \sum_{n=1}^{N_1} \log \left( \sum_{k=1}^K p_1(x_1^n \mid k) \cdot p_2(x_2^n \mid k) \cdot p_3(x_3^n \mid k) \cdot \pi_k \right)
\end{align}

The posterior distribution is:

\begin{align}
 p_{\boldsymbol{\theta}}(\mathbf{z} \mid \mathbf{D_1}) = \prod_{n=1}^{N_1} \frac{p_1(x_1^n \mid z^n) \cdot p_2(x_2^n \mid z^n) \cdot p_3(x_3^n \mid z^n) \cdot \pi_{z^n}}{\sum_{k=1}^K p_1(x_1^n \mid k) \cdot p_2(x_2^n \mid k) \cdot p_3(x_3^n \mid k) \cdot \pi_k}   
\end{align}

Similarly, for dataset $\mathbf{D_2}$:

\begin{align}
p_{\boldsymbol{\theta}}(\mathbf{D_2}, \mathbf{z'}) &= \prod_{n=1}^{N_2} p_3(x_3'^n \mid z'^n) \cdot p_4(x_4^n \mid z'^n) \cdot \pi_{z'^n} \\
\log p_{\boldsymbol{\theta}}(\mathbf{D_2}) &= \sum_{n=1}^{N_2} \log \left( \sum_{k=1}^K p_3(x_3'^n \mid k) \cdot p_4(x_4^n \mid k) \cdot \pi_k \right) \\
p_{\boldsymbol{\theta}}(\mathbf{z'} \mid \mathbf{D_2}) &= \prod_{n=1}^{N_2} \frac{p_3(x_3'^n \mid z'^n) \cdot p_4(x_4^n \mid z'^n) \cdot \pi_{z'^n}}{\sum_{k=1}^K p_3(x_3'^n \mid k) \cdot p_4(x_4^n \mid k) \cdot \pi_k}
\end{align}

\subsubsection{Method Outline}

For dataset $\mathbf{D_1}$ with latents $\mathbf{z} = \{z^n\}_{n=1}^{N_1}$, a Latent Variable Model (LVM) is defined as $p_{\boldsymbol{\theta}}(\mathbf{D_1}, \mathbf{z})$. Similarly, for $\mathbf{D_2}$ with latents $\mathbf{z'} = \{z'^n\}_{n=1}^{N_2}$, the LVM is $p_{\boldsymbol{\theta}}(\mathbf{D_2}, \mathbf{z'})$. Under independence assumptions, the distributions factorize:

\begin{align} 
p_{\boldsymbol{\theta}}(\mathbf{D_1}, \mathbf{D_2}, \mathbf{z}, \mathbf{z'}) &= p_{\boldsymbol{\theta}}(\mathbf{D_1}, \mathbf{z}) \cdot p_{\boldsymbol{\theta}}(\mathbf{D_2}, \mathbf{z'}) \\
\log p_{\boldsymbol{\theta}}(\mathbf{D_1}, \mathbf{D_2}) &= \log p_{\boldsymbol{\theta}}(\mathbf{D_1}) + \log p_{\boldsymbol{\theta}}(\mathbf{D_2}) \\
p_{\boldsymbol{\theta}}(\mathbf{z}, \mathbf{z'} \mid \mathbf{D_1}, \mathbf{D_2}) &= p_{\boldsymbol{\theta}}(\mathbf{z} \mid \mathbf{D_1}) \cdot p_{\boldsymbol{\theta}}(\mathbf{z'} \mid \mathbf{D_2})
\end{align}

To estimate ${\boldsymbol{\theta}}$, we apply the EM algorithm to maximize the marginal log-likelihoods $\log p_{\boldsymbol{\theta}}(\mathbf{D_1})$ and $\log p_{\boldsymbol{\theta}}(\mathbf{D_2})$ under latent variables. The lower bounds are given by: 

\begin{align}
\log p_{\boldsymbol{\theta}}(\mathbf{D_1}) \geq L_{D_1}(\theta, q_1), \quad \log p_{\boldsymbol{\theta}}(\mathbf{D_2}) \geq L_{D_2}(\theta, q_2)
\end{align}

where $q_1(z) = q(z \mid D_1)$ and $q_2(z) = q(z \mid D_2)$ are distributions over $Z$.

The EM algorithm steps are as follows, also detailed in Algorithm \ref{alg:2}.

\begin{algorithm}[h]
\caption{EM Algorithm}\label{alg:2}
\begin{algorithmic}[1]
\State Initialize $t = 0$ and $\boldsymbol{\theta}^{(0)} = \{\theta_1^{(0)}, \theta_2^{(0)}, \theta_3^{(0)}, \theta_4^{(0)}, \theta_Z^{(0)}\}$
\State $t \leftarrow 1$
\While{$\boldsymbol{\theta}$ not converged}
    \For{$n = 1,...,N_1$, $k = 1,...,K$}
        \State Set $q_1^{(t)}(z^n = k)$ using (\ref{eq:estep1})
    \EndFor
    \For{$n = 1,...,N_2$, $k = 1,...,K$}
        \State Set $q_2^{(t)}(z'^n = k)$ using (\ref{eq:estep2})
    \EndFor
    \State Update $\boldsymbol{\theta}^{(t)} = \{\theta_1^{(t)}, \theta_2^{(t)}, \theta_3^{(t)}, \theta_4^{(t)}, \theta_Z^{(t)}\}$ using (\ref{eq:mstep-p1}), (\ref{eq:mstep-p2}),  (\ref{eq:mstep-p3}), (\ref{eq:mstep-p4}), (\ref{eq:mstep-z})
    \State $t \leftarrow t + 1$
\EndWhile
\end{algorithmic}
\end{algorithm}

\begin{itemize}
    \item M-step: Maximize the lower bounds with respect to $\theta_1, \theta_2, \theta_3, \theta_4, \theta_Z$:
    \begin{itemize}
        \item Maximize $L_{D_1}(\theta, q_1)$ for $\theta_1, \theta_2$
        \item Maximize $L_{D_2}(\theta, q_2)$ for $\theta_4$
        \item Maximize the sum over terms containing $\theta_3$ and $\theta_Z$ across $L_{D_1}$ and $L_{D_2}$
    \end{itemize}
    \item E-step: Find $q$ to optimize $L_{D_1}(\theta, q_1) + L_{D_2}(\theta, q_2)$:
    \begin{itemize}
        \item Set $q_1$ to optimize $L_{D_1}$ given fixed $\theta$
        \item Set $q_2$ to optimize $L_{D_2}$ given fixed $\theta$
    \end{itemize}
\end{itemize}

\subsection{Deriving Algorithm Steps}

\subsubsection{Lower bound on the Likelihood}

We lower bound the log-likelihood of the observed variables:

\begin{align}
\log \left( p_{\boldsymbol{\theta}}(\mathbf{D_1}) \right) + \log \left( p_{\boldsymbol{\theta}}(\mathbf{D_2}) \right)
\end{align}

Using $q_1(z) = q(z \mid \mathbf{D_1})$ and $q_2(z) = q(z \mid \mathbf{D_2})$, the KL divergence for $\mathbf{D_1}$ is:

\begin{align}
\KL(q_1(Z) \,\|\, p_{\boldsymbol{\theta}}(Z \mid \mathbf{D_1})) = \mathbb{E}_{Z \sim q_1} \left[ \log \frac{q_1(Z)}{p_{\boldsymbol{\theta}}(Z \mid \mathbf{D_1})} \right] \geq 0
\end{align}

Thus, we have:

\begin{align}  
\log \left( p_{\boldsymbol{\theta}}(\mathbf{D_1}) \right) \geq \mathbb{E}_{Z \sim q_1} \left[ \log \frac{p_{\boldsymbol{\theta}}(\mathbf{D_1}, Z)}{q_1(Z)} \right] = L_{D_1}(\boldsymbol{\theta}, q_1)
\end{align}

where

\begin{align}
L_{D_1}(\boldsymbol{\theta}, q_1) = \sum_{n=1}^{N_1} \sum_{k=1}^{K} q_1(z^n = k) \left[ \sum_{i=1}^{3} \log p_i(x_i^n \mid k) + \log \pi_k \right] - H(q_1)
\end{align}

Similarly, for $\mathbf{D_2}$:

\begin{align}  
\log \left( p_{\boldsymbol{\theta}}(\mathbf{D_2}) \right) \geq L_{D_2}(\boldsymbol{\theta}, q_2)
\end{align}

with

\begin{align}
L_{D_2}(\boldsymbol{\theta}, q_2) = \sum_{n=1}^{N_2} \sum_{k=1}^{K} q_2({z'}^n = k) \left[ \log p_3({x'}_3^n \mid k) + \log p_4(x_4^n \mid k) + \log \pi_k \right] - H(q_2)
\end{align}

Overall, the lower bound is:

\begin{align}
\log \left( p_{\boldsymbol{\theta}}(\mathbf{D_1}) \right) + \log \left( p_{\boldsymbol{\theta}}(\mathbf{D_2}) \right) \geq L_{D_1}(\boldsymbol{\theta}, q_1) + L_{D_2}(\boldsymbol{\theta}, q_2)
\end{align}

\subsubsection{E step}

The E-step 1 updates $q_1^{(t)}(z^n = k)$ by maximizing the lower bound $L_{D_1}(\theta, q_1)$ with respect to $q_1(\cdot)$, while keeping $\theta$ fixed:

\begin{align}
q_1^{(t)}(z^n = k) &= p_{\boldsymbol{\theta}^{(t-1)}}(z^n = k \mid x_1^n, x_2^n, x_3^n) \nonumber \\
&= \frac{p_1^{(t-1)}(x_1^n \mid k) \cdot p_2^{(t-1)}(x_2^n \mid k) \cdot p_3^{(t-1)}(x_3^n \mid k) \cdot \pi_{k}^{(t-1)}}
{\sum_{j=1}^K p_1^{(t-1)}(x_1^n \mid j) \cdot p_2^{(t-1)}(x_2^n \mid j) \cdot p_3^{(t-1)}(x_3^n \mid j) \cdot \pi_{j}^{(t-1)}} \label{eq:estep1}
\end{align}

The E-step 2 updates $q_2^{(t)}({z'}^n = k)$ by maximizing $L_{D_2}(\theta, q_2)$ with respect to $q_2(\cdot)$, while keeping $\theta$ fixed:

\begin{align}
q_2^{(t)}({z'}^n = k) &= p_{\boldsymbol{\theta}^{(t-1)}}({z'}^n = k \mid {x'_3}^n, x_4^n) \nonumber \\
&= \frac{p_3^{(t-1)}({x'_3}^n \mid k) \cdot p_4^{(t-1)}(x_4^n \mid k) \cdot \pi_{k}^{(t-1)}}
{\sum_{j=1}^K p_3^{(t-1)}({x'_3}^n \mid j) \cdot p_4^{(t-1)}(x_4^n \mid j) \cdot \pi_{j}^{(t-1)}} \label{eq:estep2}
\end{align}

\subsubsection{M step: Optimal \texorpdfstring{$\theta_Z$}{}}

For the M-step, we maximize \(L_{D_1}(\theta, q_1) + L_{D_2}(\theta, q_2)\) with respect to \(\theta\), while keeping \(q(\cdot)\) fixed.

To account for the constraint \(\sum_{k=1}^K \pi_k = 1\), we use a Lagrange multiplier \(\lambda\). For any \(c \in \{1, \ldots, K\}\), we have:

\begin{align}
&\triangledown_{\pi_c} \left( L_{D_1}(\boldsymbol{\theta}, q_1) + L_{D_2}(\boldsymbol{\theta}, q_2) - \lambda \left( \sum_{k=1}^K \pi_k - 1 \right) \right) = 0 \\
&\implies \frac{\sum_{n=1}^{N_1} q_1({z}^n = c) + \sum_{n=1}^{N_2} q_2({z'}^n = c)}{\pi_c} - \lambda = 0 \\
&\implies \pi_c \propto \sum_{n=1}^{N_1} q_1({z}^n = c) + \sum_{n=1}^{N_2} q_2({z'}^n = c)
\end{align}

Since \(\sum_{k=1}^K \left( \sum_{n=1}^{N_1} q_1({z}^n = k) + \sum_{n=1}^{N_2} q_2({z'}^n = k) \right) = N_1 + N_2\), we obtain:

\begin{align}
\pi_c^{(t)} = \frac{\sum_{n=1}^{N_1} q_1^{(t)}({z}^n = c) + \sum_{n=1}^{N_2} q_2^{(t)}({z'}^n = c)}{N_1 + N_2} \label{eq:mstep-z}
\end{align}

\subsubsection{M step: Optimal \texorpdfstring{$\theta_1$}{}, \texorpdfstring{$\theta_2$}{} and \texorpdfstring{$\theta_4$}{}}

In the M-step, we use Lagrange multipliers \(\lambda(c)\) to maximize \(L_{D_1}(\boldsymbol{\theta}, q_1)\) with respect to \(p_1(m|c)\). For \(c \in \{1, \ldots, K\}\) and \(m \in \{1, \ldots, M_1\}\), we have:

\begin{align}
&\triangledown_{p_1(m|c)} \left( L_{D_1}(\boldsymbol{\theta}, q_1) - \sum_{k=1}^K \lambda(k) \left( \sum_{j=1}^{M_1} p_1(j|k) - 1 \right) \right) = 0 \\
&\implies \sum_{n=1}^{N_1} \frac{\mathbbm{1}(x_1^n = m) q_1(z^n = c)}{p_1(m|c)} - \lambda(c) = 0 \\
&\implies p_1(m|c) \propto \sum_{n=1}^{N_1} \mathbbm{1}(x_1^n = m) q_1(z^n = c)
\end{align}

Normalizing gives:

\begin{align}
p_1^{(t)}(m|c) = \frac{\sum_{n=1}^{N_1} \mathbbm{1}(x_1^n = m) q_1^{(t)}(z^n = c)}{\sum_{j=1}^{M_1} \sum_{n=1}^{N_1} \mathbbm{1}(x_1^n = j) q_1^{(t)}(z^n = c)} \label{eq:mstep-p1}
\end{align}

For \(p_2(m|c)\):

\begin{align}
p_2^{(t)}(m|c) = \frac{\sum_{n=1}^{N_2} \mathbbm{1}(x_2^n = m) q_1^{(t)}(z^n = c)}{\sum_{j=1}^{M_2} \sum_{n=1}^{N_2} \mathbbm{1}(x_2^n = j) q_1^{(t)}(z^n = c)} \label{eq:mstep-p2}
\end{align}

Similarly, for \(p_4(m|c)\), we maximise \(L_{D_2}(\boldsymbol{\theta}, q_2)\):

\begin{align}
p_4^{(t)}(m|c) = \frac{\sum_{n=1}^{N_2} \mathbbm{1}(x_4^n = m) q_2^{(t)}(z'^n = c)}{\sum_{j=1}^{M_4} \sum_{n=1}^{N_2} \mathbbm{1}(x_4^n = j) q_2^{(t)}(z'^n = c)} \label{eq:mstep-p4}
\end{align}

\subsubsection{M step: Optimal \texorpdfstring{$\theta_3$}{}}

In the M-step, we use Lagrange multipliers \(\lambda(c)\) to maximize \(L_{D_1}(\boldsymbol{\theta}, q_1)\) + \(L_{D_2}(\boldsymbol{\theta}, q_2)\) with respect to \(p_3(m|c)\). For \(c \in \{1, \ldots, K\}\) and \(m \in \{1, \ldots, M_3\}\), we have:

\begin{align}
&\triangledown_{{p_3}(m|c)} \left( L_{D_1}(\boldsymbol{\theta}, q_1) + L_{D_2}(\boldsymbol{\theta}, q_2) - \sum_{k=1}^K \lambda(k)(\sum_{j=1}^{M_2} {p_3}(j | k)  - 1) \right) = 0 \\
&\implies \frac{ \sum_{n=1}^{N_1} \mathbbm{1}({x_3}^n = m) q_1({z}^n = c) + \sum_{n=1}^{N_2} \mathbbm{1}({x'_{2}}^n = m) q_2({z'}^n = c)}{{p_3}(m|c)}  -  \lambda(k = c) = 0 \\ 
&\implies {p_3}(m|c) \propto \sum_{n=1}^{N_1} \mathbbm{1}({x_3}^n = m) q_1({z}^n = c) + \sum_{n=1}^{N_2} \mathbbm{1}({x'_{2}}^n = m) q_2({z'}^n = c) 
\end{align}

We therefore obtain M step update 

\begin{align}
{p_3^{(t)}}(m | c) = \frac{\sum_{n=1}^{N_1} \mathbbm{1}({x_3}^n = m) q_1^{(t)}({z}^n = c) + \sum_{n=1}^{N_2}  \mathbbm{1}({x'_3}^n = m) q_2^{(t)}({z'}^n = c)}{ \sum_{j=1}^{M_2} \left( \sum_{n=1}^{N_1} \mathbbm{1}({x_3}^n = j) q_1^{(t)}({z}^n = c) + \sum_{n=1}^{N_2}  \mathbbm{1}({x'_3}^n = j) q_2^{(t)}({z'}^n = c) \right)}  \label{eq:mstep-p3}
\end{align}

\section{Dataset Details}
\label{appendix:datasets}
For the Adult Data, the `fnlwgt' attribute is dropped as it is not relevant to the task and the `education-num' attribute as it duplicates the information available in the `education' attribute. COMPAS Data is filtered to only include `race` column is either `African-American' or `Caucasian' and coding as $\{black, white\}$. We further combine three columns containing juvenile crime counts to get the total number of juvenile crimes. Details of the attributes and their values can be found in Tables \ref{tab:adult}, \ref{tab:compas}, and \ref{tab:german_credit}.

\begin{table}[h]
  \caption{Adult Data: Attributes and Their Values}
  \label{tab:adult}
  \centering
  \begin{tabular}{lp{10cm}}
    \toprule
    \textbf{Attribute} & \textbf{Values} \\
    \midrule
    Age & \{25--60, \textless 25, \textgreater 60\} \\
    Capital Gain & \{\textless=5000, \textgreater 5000\} \\
    Capital Loss & \{\textless=40, \textgreater 40\} \\
    Education & \{assoc-acdm, assoc-voc, bachelors, doctorate, HS-grad, masters, prof-school, some-college, high-school, primary/middle school\} \\
    Hours Per Week & \{\textless 40, 40--60, \textgreater 60\} \\
    Income & \{\textless=50K, \textgreater 50K\} \\
    Marital Status & \{married, other\} \\
    Native Country & \{US, non-US\} \\
    Occupation & \{adm-clerical, armed-forces, craft-repair, exec-managerial, farming-fishing, handlers-cleaners, machine-op-inspct, other-service, priv-house-serv, prof-specialty, protective-serv, sales, tech-support, transport-moving\} \\
    Race & \{non-white, white\} \\
    Relationship & \{non-spouse, spouse\} \\
    Sex & \{male, female\} \\
    Workclass & \{private, non-private\} \\
    \bottomrule
  \end{tabular}
\end{table}

\begin{table}[h]
  \caption{COMPAS Data: Attributes and Their Values}
  \label{tab:compas}
  \centering
  \begin{tabular}{lp{10cm}}
    \toprule
    \textbf{Attribute} & \textbf{Values} \\
    \midrule
    Age Category & \{25 - 45, \textgreater 45, \textless 25\} \\
    Charge Degree & \{F, M\} \\
    Juvenile Crime & \{0, 1, 2, 3, 4, 5, 6, 7, 8, 9, 10, 11, 14\} \\
    Priors Count & \{0, 1, 2, 3, 4, 5, 6, 7, 8, 9, 10, 11, 12, 13, 14, 15, 16, 17, 18, 19, 20, 21, 22, 23, 24, 25, 26, 27, 28, 29, 30, 31, 33, 36, 37, 38\} \\
    Race & \{Black, White\} \\
    Score Text & \{High, Low, Medium\} \\
    Sex & \{Female, Male\} \\
    Two-Year Recidivism & \{0, 1\} \\
    Violent Score Text & \{High, Low, Medium\} \\
    \bottomrule
  \end{tabular}
\end{table}

\begin{table}[h]
  \caption{German Credit Data: Attributes and Their Values}
  \label{tab:german_credit}
  \centering
  \begin{tabular}{p{3cm}p{10cm}}
    \toprule
    \textbf{Attribute} & \textbf{Values} \\
    \midrule
    Age & \{<= 25, >25\} \\
    Checking Account & \{0 <= <200 DM, <0 DM, >= 200 DM, no account\} \\
    Class Label & \{0, 1\} \\
    Credit Amount & \{<=2000, 2001-5000, >5000\} \\
    Credit History & \{all credits at this bank paid back duly, critical account, delay in paying off, existing credits paid back duly till now, no credits taken\} \\
    Duration & \{<=6, 7-12, >12\} \\
    Employment Since & \{1 <= < 4 years, 4 <= <7 years, <1 years, >=7 years, unemployed\} \\
    Existing Credits & \{1, 2, 3, 4\} \\
    Foreign Worker & \{no, yes\} \\
    Housing & \{for free, own, rent\} \\
    Installment Rate & \{1, 2, 3, 4\} \\
    Job & \{management/ highly qualified employee, skilled employee / official, unemployed/ unskilled - non-resident, unskilled - resident\} \\
    Marital Status & \{divorced/separated, married/widowed\} \\
    Number of People Provide Maintenance For & \{1, 2\} \\
    Other Debtors & \{co-applicant, guarantor, none\} \\
    Other Installment Plans & \{bank, none, store\} \\
    Property & \{car or other, real estate, savings agreement/life insurance, unknown / no property\} \\
    Purpose & \{business, car (new), car (used), domestic appliances,
education, furniture/equipment, others, radio/television,
repairs, retraining\} \\
    Residence Since & \{1, 2, 3, 4\} \\
    Savings Account & \{100 <= <500 DM, 500 <= < 1000 DM, <100 DM, >= 1000 DM, no savings account\} \\
    Sex & \{female, male\} \\
    Telephone & \{none, yes\} \\
    \bottomrule
  \end{tabular}
\end{table}

\section{Evaluating Quality of Synthetic Data}
\label{appendix:quality}

\subsection{Overall Fidelity Metrics: Synthetic vs. Real Data}
\label{appendix:fidelity}

Full results for overall fidelity metrics, including Total Variation Distance Complement (1-TVD), Contingency Similarity (CS), and Discriminator Measure (DM) across various synthetic datasets, are presented in \tabref{tab:fidelity}. This table provides a comprehensive comparison of the fidelity of different synthetic data generation methods to real-world data.

\begin{figure*}[ht]
    \centering
    \subfloat[\label{fig:corr1}]
    {
        {
            \includegraphics[scale=0.2]{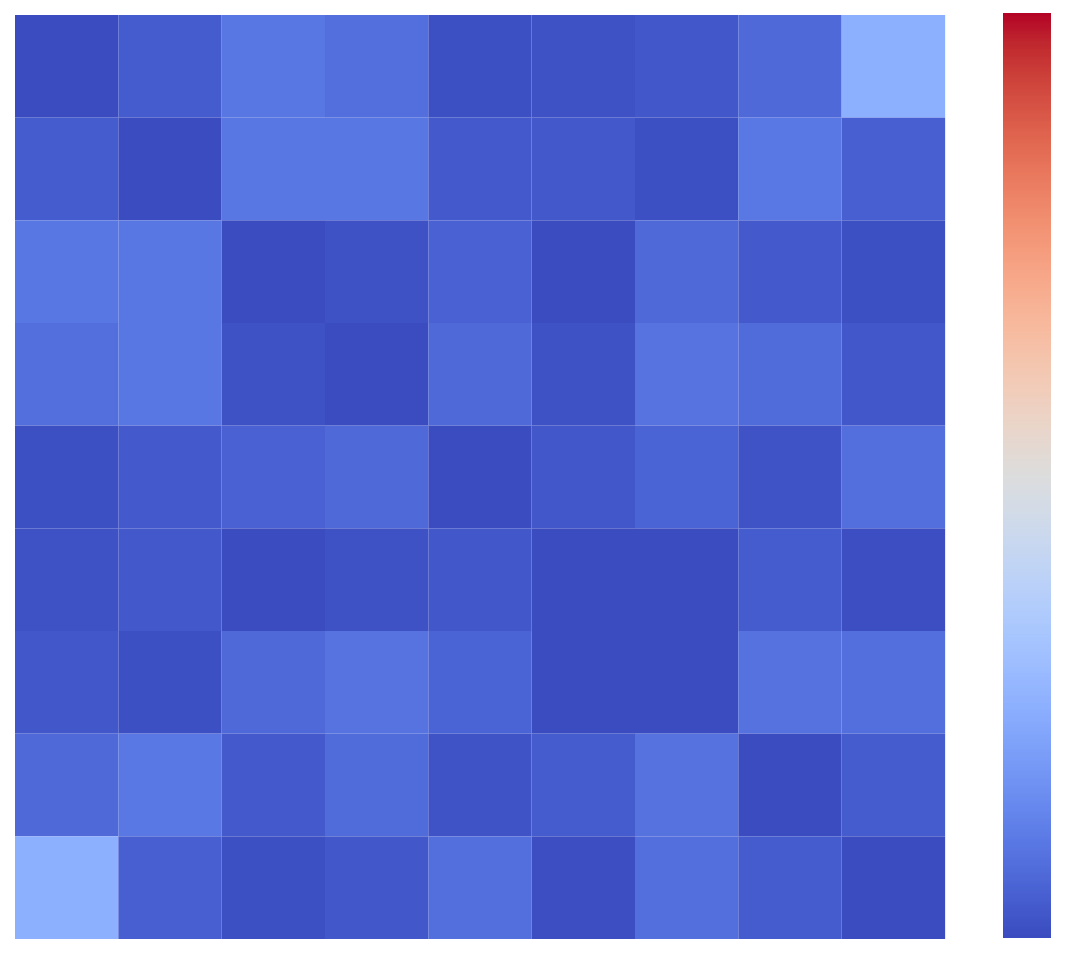}
        }
    }%
    \quad
    \subfloat[\label{fig:corr2}]
    {
        {
            \includegraphics[scale=0.2]{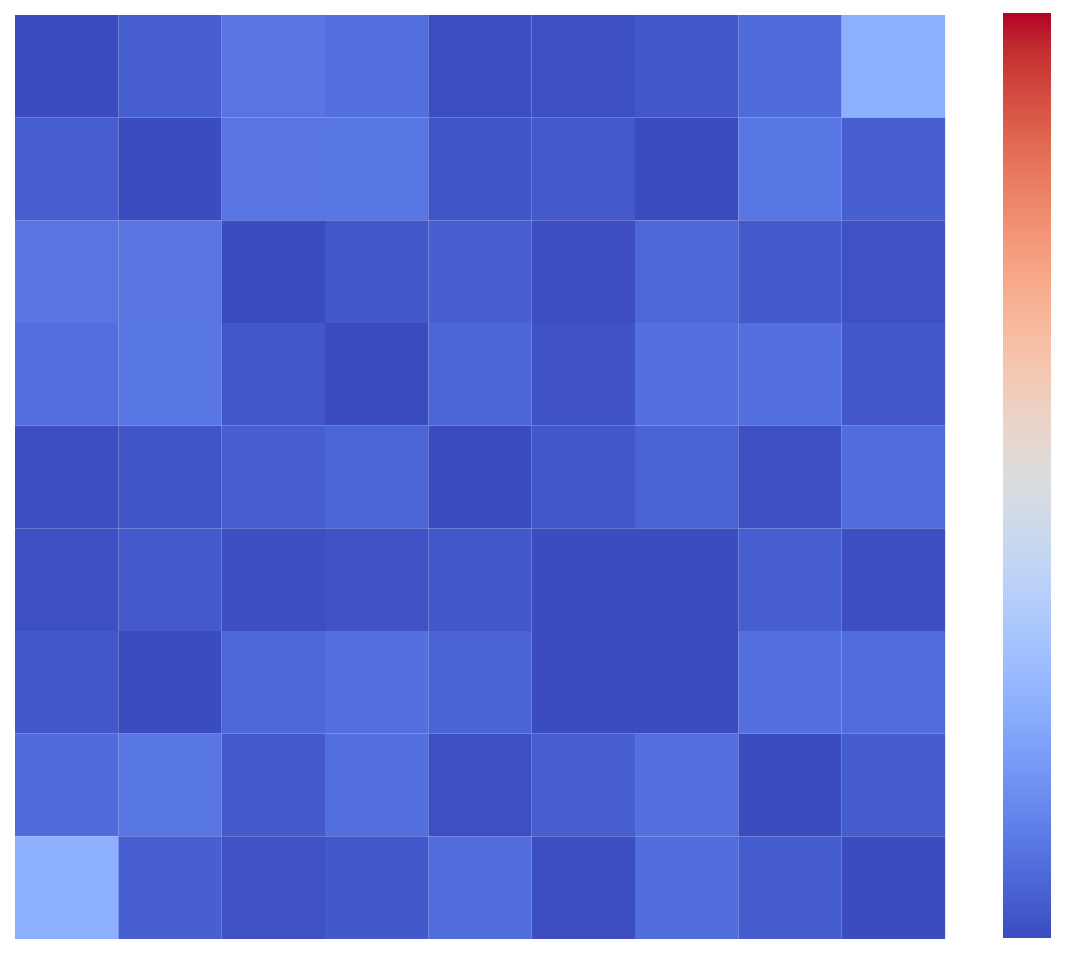}
        }
    }%
        \quad
    \subfloat[\label{fig:corr3}]
    {
        {
            \includegraphics[scale=0.2]{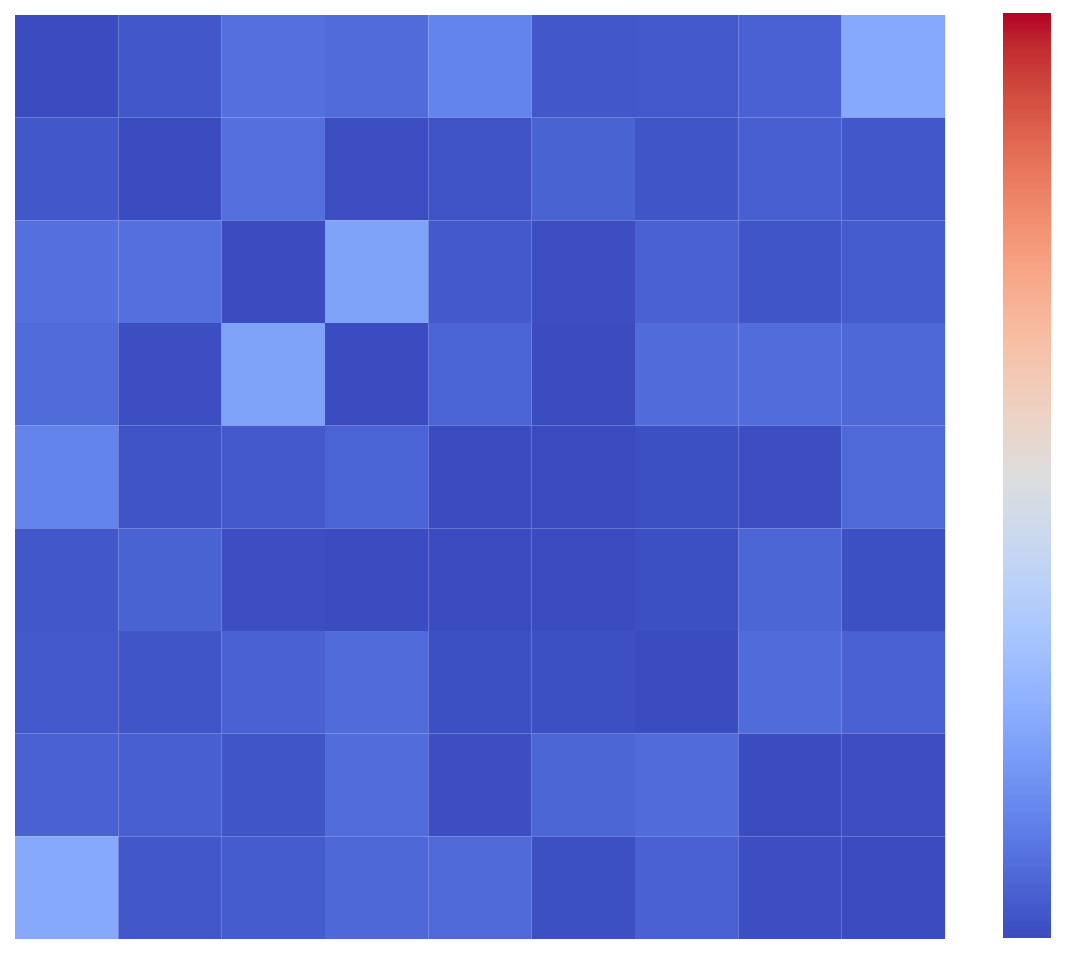}
        }
    }%
            \quad
    \subfloat[\label{fig:corr4}]
    {
        {
            \includegraphics[scale=0.2]{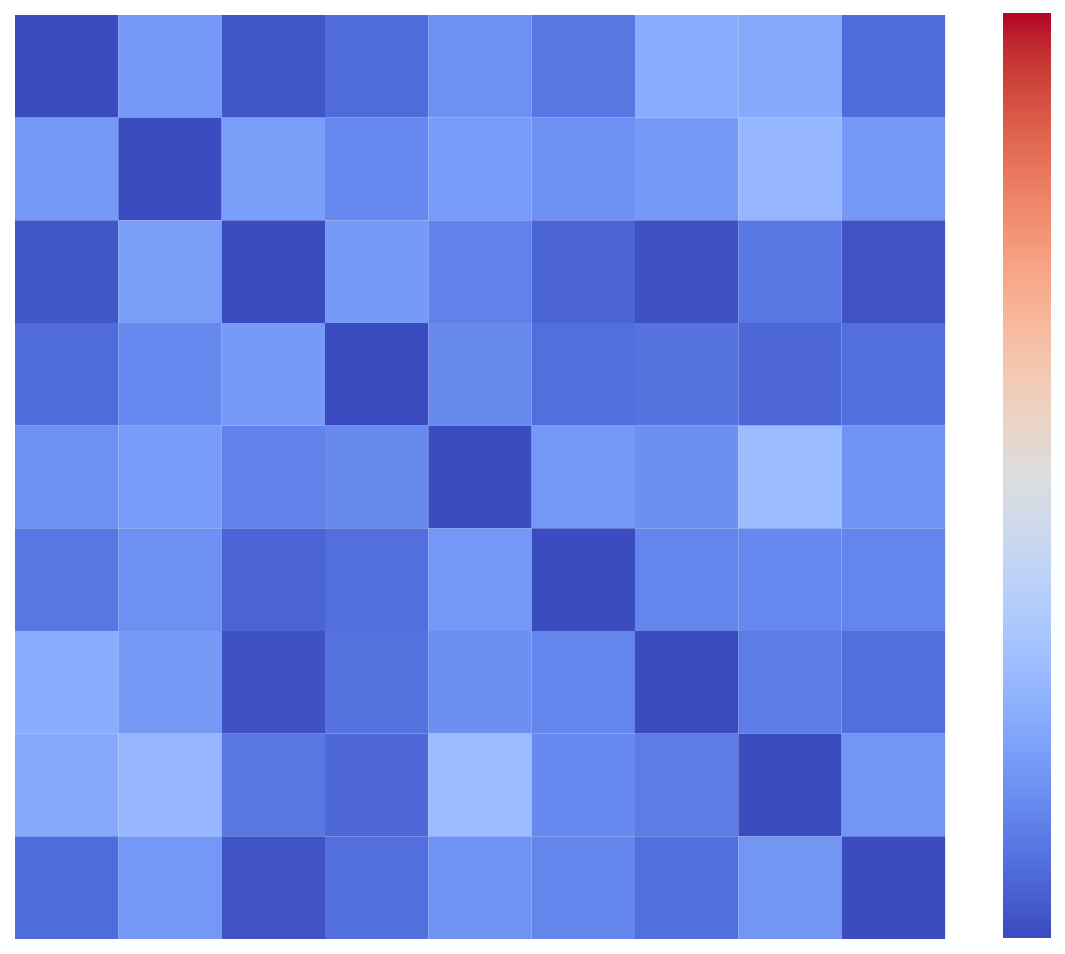}
        }
    }%
                \quad
    \subfloat[\label{fig:corr5}]
    {
        {
            \includegraphics[scale=0.2]{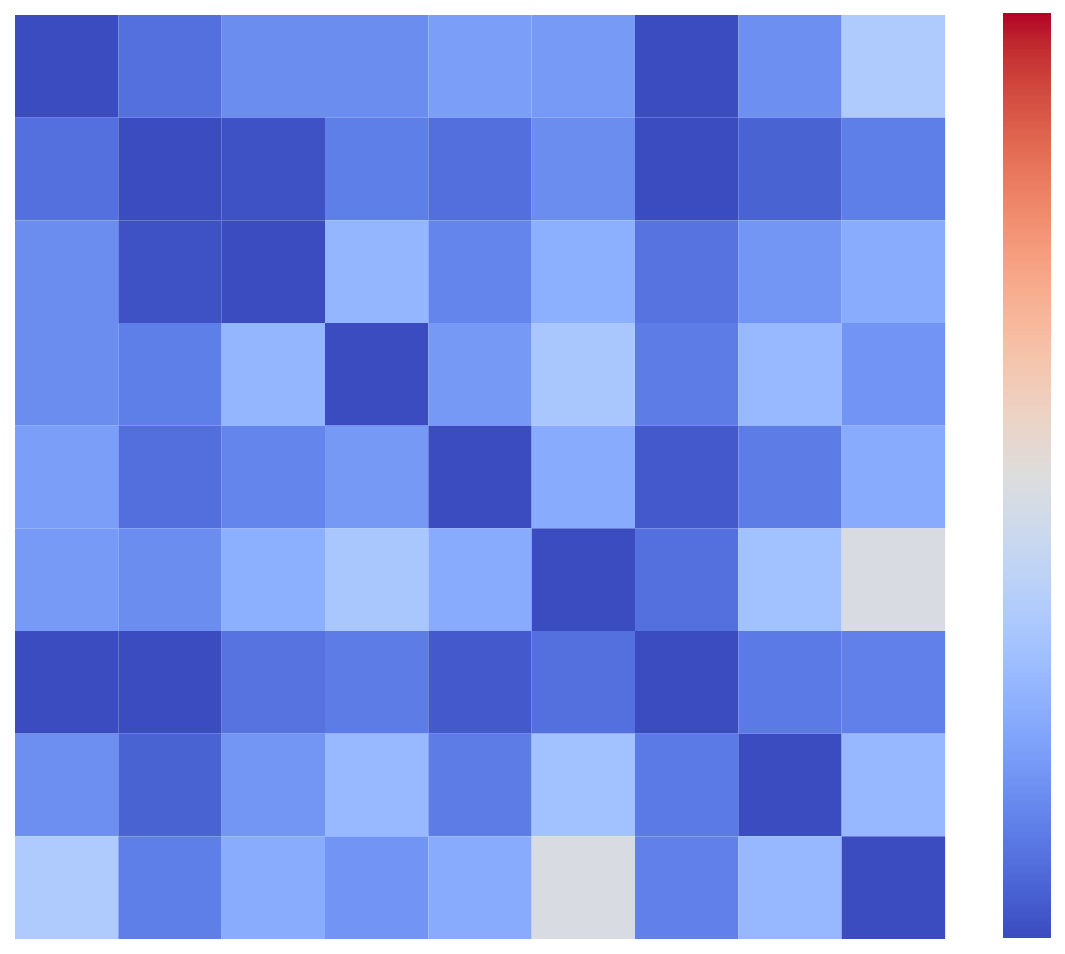}
        }
    }%
    
    \caption{Difference in Cramér's V Correlation (DCC) for pairs of attributes in synthetic test data and in real test data. Values close to zero (dark blue colour) indicate synthetic data is more similar to real data. Results shown for COMPAS Data, with synthetic data  generated from separate data with overlapping variable `Score'.  Subplots correspond to different joint estimation methods (a) Independence given Overlap, (b) Marginal Preservation (c) Latent Naïve Bayes. (d) CTGAN Baseline (e) Independent Baseline. }
    \label{fig:correlation}
\end{figure*}

\figref{fig:correlation} shows the difference in Cramér's V correlation (DCC) between synthetic and real test data for COMPAS. Similar patterns are observed across other synthetic datasets.

\begin{table*}[ht]
\centering
\caption{Fidelity metrics for synthetic datasets generated from separate data (overlapping variable in brackets next to joint distribution estimation method). Metrics include total variation distance complement (1-TVD), contingency similarity (CS), discriminator measure (DM). Baseline methods include CTGAN and Indep.}
\begin{tabular}{llccc}
\toprule
\textbf{Dataset} & \textbf{Method (Overlapping)} & \textbf{1-TVD} $\uparrow$ & \textbf{CS} $\uparrow$ & \textbf{DM} $\downarrow$ \\ 
\midrule
\textbf{Adult} \\  
                 & Indep-Overlap (Relationship) & 0.993 & \textbf{0.983} & 0.588 \\
                 & Marginal (Relationship)      & 0.993 &\textbf{ 0.983} & 0.588 \\
                 & Latent (Relationship)        & 0.986 & 0.968 & 0.658 \\
                 & Indep-Overlap (Marital Status) & \textbf{0.994} & \textbf{0.983} & \textbf{0.587} \\
                 & Marginal (Marital Status)    & 0.993 & 0.982 & 0.594 \\
                 & Latent (Marital Status)      & 0.987 & 0.970 & 0.655 \\
                 & CTGAN                        & 0.935 & 0.938 & 0.656 \\
                 & Indep                      & 0.935 & 0.895 & 0.808 \\ 
                 \midrule
\textbf{COMPAS} \\  
                 & Indep-Overlap (Score)        & 0.978 & 0.952 & 0.596 \\
                 & Marginal (Score)             & \textbf{0.979} & 0.953 & 0.598 \\
                 & Latent (Score)               & 0.978 & 0.951 & 0.592 \\
                 & Indep-Overlap (Violent Score) & 0.978 & \textbf{0.955} & 0.577 \\
                 & Marginal (Violent Score)     & 0.978 & \textbf{0.955} & \textbf{0.573 }\\
                 & Latent (Violent Score)       & 0.976 & 0.950 & 0.598 \\ 
                & CTGAN                        & 0.910 & 0.839 & 0.699 \\
                 & Indep                       & \textbf{0.979} & 0.913 & 0.689 \\ 
                 \midrule
\textbf{German} \\  
                 & Indep-Overlap (Property)     & 0.965 & 0.926 & 0.613 \\
                 & Marginal (Property)          & \textbf{0.966} & 0.926 & 0.628 \\
                 & Latent (Property)            & 0.965 & 0.924 & 0.586 \\
                 & Indep-Overlap (Housing)      & \textbf{0.966} & 0.926 & 0.618 \\
                 & Marginal (Housing)           & \textbf{0.966} & \textbf{0.927 }& 0.621 \\
                 & Latent (Housing)             & \textbf{0.966} & 0.925 & \textbf{0.575} \\ 
                & CTGAN                        & 0.946 & 0.894 & 0.697 \\
                 & Indep                        & 0.965 & 0.920 & 0.696 \\ 
                 \bottomrule
\end{tabular}
\label{tab:fidelity}
\end{table*}

\subsection{Joint Distribution of Protected Attributes and Outcomes: Synthetic vs. Real Data}
\label{appendix:kl}

KL Divergence Values for the joint distribution of protected attributes and outcome labels between synthetic and real data, evaluated across various methods and data separations, are detailed in \tabref{tab:kl-divergence-table}.

\begin{table}[h]
  \caption{KL divergence of $p(A,Y)$ in synthetic vs real data (where Y is the outcome label and A is a protected attribute such as race, sex, and age). Synthetic datasets generated from separate data (with the overlapping variable indicated in brackets next to the joint distribution estimation method). Baseline methods include CTGAN and Indep.}
  \label{tab:kl-divergence-table}
  \centering
  \begin{tabular}{llccc}
    \toprule
    \textbf{Dataset} & \textbf{Method (Overlapping)} & \textbf{KL for Race} $\downarrow$ & \textbf{KL for Sex} $\downarrow$  & \textbf{KL for Age} $\downarrow$  \\
    \midrule
    \textbf{Adult} \\
    & Indep-Overlap (Relationship) & \textbf{0.002} & \textbf{0.001} & -- \\
    & Marginal (Relationship)      & \textbf{0.002} & \textbf{0.001} & -- \\
    & Latent (Relationship)        & \textbf{0.002} & 0.002 & -- \\
    & Indep-Overlap (Marital Status) & \textbf{0.002} & \textbf{0.001} & -- \\
    & Marginal (Marital Status)      & \textbf{0.002} & 0.002 & -- \\
    & Latent (Marital Status)        & \textbf{0.002} & \textbf{0.001} & -- \\ 
    & CTGAN                 & 0.132 & 0.048 & -- \\
    & Indep                & 0.005 & 0.026 & -- \\
    \midrule
    \textbf{COMPAS} \\
    & Indep-Overlap (Score)         & 0.006 & 0.044 & -- \\
    & Marginal (Score)              & \textbf{0.005} & 0.039 & -- \\
    & Latent (Score)                & \textbf{0.005} & 0.034 & -- \\
    & Indep-Overlap (Violent Score) & 0.015 & 0.038 & -- \\
    & Marginal (Violent Score)      & 0.015 & 0.038 & -- \\
    & Latent (Violent Score)        & \textbf{0.005} & \textbf{0.026} & -- \\ 
    & CTGAN                & 0.498 & 0.506 & -- \\
    & Indep                & 0.058 & 0.062 & -- \\
    \midrule
    \textbf{German} \\
    & Indep-Overlap (Property)      & -- & 0.015 & 0.052 \\
    & Marginal (Property)           & -- & 0.013 & 0.055 \\
    & Latent (Property)             & -- & 0.003 & 0.023 \\
    & Indep-Overlap (Housing)       & -- & 0.003 & 0.034 \\
    & Marginal (Housing)            & --& 0.005 & 0.035 \\
    & Latent (Housing)              & -- & \textbf{0.002} & \textbf{0.022 }\\ 
    & CTGAN                 & -- & 0.282 & 0.215 \\
    & Indep                 & -- & 0.007 & 0.038 \\
    \bottomrule
  \end{tabular}
\end{table}

\subsection{Detailed Fairness Metrics Comparison: Synthetic vs. Real Data}
\label{appendix:fairness-metrics}

\tabref{tab:fairness_metrics} provides a detailed comparison of absolute differences in fairness metrics for a Decision Tree classifier, as evaluated on various synthetic datasets compared to real test data. The metrics include Average Odds Difference (AOD), Disparate Impact (DI), and Equal Opportunity Difference (EOD). The analysis is based on 1000 bootstrapped samples. The table summarises these metrics across different synthetic datasets and baselines.

\begin{table}[h]
  \caption{Absolute differences between bootstrap means of fairness metrics from synthetic and real data. Metrics calculated for Decision Tree Classifier, using 1000 bootstrapped aamples. Metrics include AOD (Average Odds Difference), DI (Disparate Impact), and EOD (Equal Opportunity Difference). Synthetic datasets generated from separate data (with the overlapping variable indicated in brackets next to the joint distribution estimation method). Baseline methods include CTGAN and Indep.}
  \label{tab:fairness_metrics}
  \centering
  \resizebox{\textwidth}{!}{
  \begin{tabular}{llccccccc}
    \toprule
    \textbf{Dataset} & \textbf{Method (Overlapping)} & \multicolumn{3}{c}{\textbf{Race}} & \multicolumn{3}{c}{\textbf{Sex}}  \\
    \cmidrule(lr){3-5} \cmidrule(lr){6-8} 
    && \textbf{AOD} $\downarrow$ & \textbf{DI} $\downarrow$ & \textbf{EOD} $\downarrow$ & \textbf{AOD} $\downarrow$ & \textbf{DI} $\downarrow$ & \textbf{EOD} $\downarrow$  \\
    \midrule
    \textbf{Adult} \\
    &Indep-Overlap (Marital) & 0.008 & 0.137 & 0.015 & 0.015 & 0.030 & 0.038 \\
    &Indep-Overlap (Relationship) & 0.013 & 0.112 & 0.022 & 0.037 & 0.006 & 0.070 \\
    &Latent (Marital) & 0.013 & 0.063 & 0.016 & 0.069 & 0.162 & 0.134 \\
    &Latent (Relationship) & \textbf{0.003} & 0.079 & 0.010 & 0.068 & 0.154 & 0.148 \\
    &Marginal (Marital) & \textbf{0.003} & 0.144 & \textbf{0.005} & \textbf{0.010} & 0.047 & \textbf{0.030} \\
    &Marginal (Relationship) & 0.009 & 0.112 & 0.014 & 0.036 & \textbf{0.002} & 0.071 \\ 
    &CTGAN & 0.027 & \textbf{0.003 }& 0.053 & 0.013 & 0.048 & 0.055 \\
    &Indep  & 0.021 & 0.390 & 0.023 & 0.070 & 0.669 & 0.071 \\
    \midrule
    \textbf{COMPAS} \\
    &Indep-Overlap (Score) & 0.002 & 0.003 & 0.034 & 0.040 & 0.079 & 0.014 \\
    &Indep-Overlap (Violent Score) & 0.031 & 0.046 & 0.057 & 0.013 & 0.032 & 0.027 \\
    &Latent (Score) & 0.001 & 0.001 & 0.035 &  0.013 & 0.032 & 0.030 \\
    &Latent (Violent Score) & 0.005 & 0.010 & \textbf{0.016} & \textbf{0.009} & \textbf{0.016} & 0.054 \\
    &Marginal (Score) & \textbf{0.000} & \textbf{0.000} & 0.037  & 0.029 & 0.061 & \textbf{0.010} \\
    &Marginal (Violent Score) & 0.039 & 0.057 & 0.063 & 0.015 & 0.037 & 0.034  \\ 
    &CTGAN & 0.065 & 0.212 & 0.097 & 0.083 & 0.351 & 0.135 \\
    &Indep & 0.134 & 0.211 & 0.146 & 0.072 & 0.138 & 0.012 \\
    \midrule
     &  & \multicolumn{3}{c}{\textbf{Age}} & \multicolumn{3}{c}{\textbf{Sex}}  \\
    \cmidrule(lr){3-5} \cmidrule(lr){6-8} 
    && \textbf{AOD} $\downarrow$ & \textbf{DI}  $\downarrow$& \textbf{EOD} $\downarrow$ & \textbf{AOD} $\downarrow$ & \textbf{DI} $\downarrow$ & \textbf{EOD} $\downarrow$  \\
    \cmidrule{3-8}
    \textbf{German} \\
    &Indep-Overlap (Housing) & \textbf{0.106} & 0.178 & 0.084 & 0.048 & 0.068 & 0.071 \\
    &Indep-Overlap (Property) & 0.144 & 0.216 & 0.085 & 0.010 & 0.020 & 0.048 \\
    &Latent (Housing) & 0.141 & 0.190 & 0.049 & 0.025 & 0.049 & 0.069 \\
    &Latent (Property) & 0.119 & \textbf{0.170} & \textbf{0.048} & 0.014 & 0.034 & 0.062 \\
    &Marginal (Housing) & 0.111 & 0.178 & 0.076 & 0.044 & 0.058 & 0.055 \\
    &Marginal (Property) & 0.150 & 0.218 & 0.078 & \textbf{0.005} & \textbf{0.015 }& \textbf{0.045} \\ 
    &CTGAN & 0.140 & 0.199 & 0.066 & 0.009 & 0.019 & 0.048 \\
    &Indep & 0.139 & 0.195 & 0.062 & 0.007 & 0.024 & 0.056 \\
    \bottomrule
  \end{tabular}
  }
\end{table}


\end{document}